%% file: main.tex
\documentclass{article}

\input{set_up}
\usepackage{PRIMEarxiv}
\usepackage{footmisc} 
\usepackage[utf8]{inputenc} 
\usepackage[T1]{fontenc}    
\usepackage{hyperref}       
\usepackage{url}            
\usepackage{booktabs}       
\usepackage{amsfonts}       
\usepackage{nicefrac}       
\usepackage{microtype}      
\usepackage{lipsum}
\usepackage{fancyhdr}       
\usepackage{graphicx}       
\graphicspath{{media/}}     

\title{Generalization Ability of Wide Residual Networks}

\author{Jianfa Lai\footnotemark[1] , \\
Center for Statistical Science, Department of Industrial Engineering \\ Tsinghua University, Beijing, China \\
\texttt{Jianfalai@mail.tsinghua.edu.cn, } \\
\And
 Zixiong Yu\footnotemark[1]~, Songtao Tian \\
Department of Mathematical Sciences  \\ Tsinghua University, Beijing, China \\
\texttt{\{yuzx19,tst20\}@mails.tsinghua.edu.cn}
\And
Qian Lin\footnotemark[2] \footnotemark[3]\\
Center for Statistical Science, Department of Industrial Engineering \\ Tsinghua University, Beijing, China \\
\texttt{qianlin@tsinghua.edu.cn}
}

\begin{document}
\maketitle

\renewcommand{\thefootnote}{\fnsymbol{footnote}} 
\footnotetext[1]{These authors contributed equally to this work.} 
\footnotetext[2]{Corresponding author.} 
\footnotetext[3]{Qian Lin also affiliates with Beijing Academy of Artificial Intelligence, Beijing, China.} 

\begin{abstract}
In this paper, we study the generalization ability of the wide residual network on $\mathbb{S}^{d-1}$ with the ReLU activation function.
We first show that as the width $m\rightarrow\infty$, the residual network kernel (RNK) uniformly converges to the  residual neural tangent kernel (RNTK). This uniform convergence further  guarantees that the generalization error of the residual network converges to that of the kernel regression with respect to the RNTK. 
As direct corollaries,  we then show $i)$ the wide residual network with the early stopping strategy can achieve the minimax rate provided that the target regression function falls in the reproducing kernel Hilbert space (RKHS) associated with the RNTK;
$ii)$ the wide residual network can not generalize well if it is trained till overfitting the data. We finally illustrate some experiments to reconcile the contradiction between our theoretical result and the widely observed ``benign overfitting phenomenon''
\end{abstract}


\section{Introduction}
\input{s1_introduction}

\section{Residual neural tangent kernel}

\input{s2_NTK}

\section{Uniform convergence of wide residual networks}\label{sec:main_result}
\input{s3_uniform_convergence}

\section{Generalization ability of wide residual networks}\label{sec:generalization}
\input{s4_generalization}

\section{Experiments}\label{sec:experiment}
\input{s5_experiment}

\section{Conclusion and discussion}

\input{s6_conclusion}


\bibliographystyle{plain}
\bibliography{main.bib}

\input{sa_Proof_uniform_convergence}
\input{sb_Proof_generalization}

\end{document}

%% file: set_up.tex
\PassOptionsToPackage{numbers, compress}{natbib}

\usepackage[utf8]{inputenc} 
\usepackage[T1]{fontenc}    
\usepackage{hyperref}       
\usepackage{url}            
\usepackage{booktabs}       
\usepackage{amsfonts}       
\usepackage{nicefrac}       
\usepackage{microtype}      
\usepackage{xcolor}         
\usepackage{mleftright}

\RequirePackage{graphicx}
\RequirePackage{amsthm,amsmath,amsfonts,amssymb}
\RequirePackage[numbers]{natbib}
\newcommand{\pt}[1]{\left(#1\right)}
\newcommand{\bk}[1]{\left[#1\right]}
\newcommand{\cl}[1]{\left\{#1\right\}}
\newcommand{\ag}[1]{\left\langle#1\right\rangle}
\newcommand{\fl}[1]{\left\lfloor#1\right\rfloor}

\newcommand{\norm}[1]{\left\lVert#1\right\rVert}
\newcommand{\abs}[1]{\left\lvert#1\right\rvert}
\newcommand{\mpt}[1]{\mleft(#1\mright)}
\newcommand{\mbk}[1]{\mleft[#1\mright]}

\newcommand{\bm}{\boldsymbol}
\newcommand{\mb}{\mathbb}
\newcommand{\mr}{\mathrm}

\newcommand{\poly}{\mr{poly}}
\newcommand{\me}{\mr{e}}

\usepackage{verbatim}
\newtheorem{theorem}{Theorem}[section]
\newtheorem{lemma}[theorem]{Lemma}
\newtheorem{corollary}[theorem]{Corollary}
\newtheorem{proposition}[theorem]{Proposition}
\newtheorem{assumption}{Assumption}

\newcommand{\widesim}{\mathrel{\scalebox{2.2}[1]{\hbox{$\sim$}}}}
\newcommand{\iid}{\mathrel{\stackrel{\mr{i.i.d.}}{\widesim}}}

\usepackage[normalem]{ulem}

%% file: s1_introduction.tex
Deep neural networks have led to many great successes in various fields. It is widely observed that the performance of the network is highly dependent on the architecture of the network \cite{krizhevsky2017imagenet,he2016deep,vaswani2017attention}. A prominent architecture, leading to a major improvement in the current deep learning, is the residual network, known also as ResNet \cite{he2016deep}.
The residual network introduced the skip connection, which makes training a network with hundreds or thousands of layers possible. Since \cite{he2016deep} invented the residue neural network, it has been widely applied in various fields and has obtained incredible success.

Many works tried to explain the success of residual networks, however, most of them focus on the optimization aspects. For example, thanks to the skip connection,
\cite{veit2016residual} showed that  residual networks can avoid the vanishing gradient problem;
\cite{li2018visualizing} showed that it is easier to train a network with skip connections since the loss landscape is smoother compared with the network without skip connections;
\cite{liu2019towards} showed that the network with skip connections can avoid spurious local minima.
However, it is still unclear how the skip connection affects the generalization ability of neural networks.

Jacot et al. \cite{jacot2018neural} made a seminal observation that the training process of wide neural networks can be well approximated by that of kernel regression with respect to the NTK as the width $m\rightarrow\infty$. 
More precisely, they first interpreted the gradient flow of neural networks as a gradient flow associated to a time-varying kernel regression problem and this time-varying kernel is called the neural network kernel (NNK). 
They then observed that  as the width $m\to\infty$, the NNK pointwisely converges to the neural tangent kernel (NTK), a time-invariant kernel  during the training process. 
Inspired by the idea of the NTK, \cite{lai2023generalization, li2023Statistical} theoretically proved that the generalization error of wide fully-connected neural networks can be approximated by that of kernel regression with respect to the NTK. In other words, one can study the generalization ability of wide neural networks by studying the generalization ability of kernel regression with respect to the NTK.

The generalization ability of kernel regression was an active field two decades ago.
With the polynomial decay of the eigenvalues associated to the kernel, \cite{blanchard2018optimal,lin2020optimal,zhang2023optimality} showed the spectral algorithms, including the early-stopped kernel regression with gradient flow, are minimax optimal under some mild conditions. 
\cite{rakhlin2019consistency,beaglehole2022kernel, buchholz2022kernel, li2023kernel} and \cite{liang2020just} considered the generalization performance of kernel ridgeless regression in low dimensional and high dimensional data respectively. 
Some papers reinvestigated the generalization performance of kernel ridge regression under the Gaussian design assumption of the eigenfunctions \cite{bordelon2020spectrum,jacot2020kernel,cui2021generalization,mallinar2022benign} and offered some elaborate results. For example, \cite{cui2021generalization} depicted the generalization error of kernel ridgeless regression under different source conditions, regularization and noise levels. 

Some researchers have analyzed the residual neural tangent kernel (RNTK) which is first introduced in 
\cite{huang2020deep}.
\cite{huang2020deep} showed that the residual network kernel (RNK) at the initialization converges to the RNTK as the width $m\rightarrow\infty$. 
\cite{tirer2022kernel} further showed the stability of RNK and the RNK point-wisely convergences to the RNTK during the training process. 
However, the ReLU function, the commonly used activation function, does not satisfy the assumptions in \cite{tirer2022kernel}. 
\cite{belfer2021spectral} showed that for the input distributed uniformly on the hypersphere $\mathbb{S}^{d-1}$, the eigenfunctions of RNTK are the spherical harmonics and the $k^{th}$ multiple eigenvalue of the RNTK $\mu_k\asymp k^{-d}$.

Though the above works offer some understanding of the residual networks and the RNTK, they say nothing about the generalization ability of the residual networks. 
In this paper, we perform a study on the generalization ability of the residual network on $\mathbb{S}^{d-1}$.
In Section \ref{sec:main_result}, we first show that the RNK converges to the RNTK uniformly during the training process as the width $m\to \infty$, 
therefore the generalization error of the residual network is well approximated by that of kernel regression with respect to the RNTK.
With this approximation, we then show in Section \ref{sec:generalization} that:
$i)$ the residual network produced by an early stopping  strategy is minimax rate optimal; $ii)$ the overfitted residual network can not generalize well. 
It is clear that the ``benign overfitting phenomenon'' found on neural networks violates the latter statement. To reconcile this contradiction, we further illustrate some experiments on the role played by the signal strength in the ``benign overfitting phenomenon'' in Section \ref{sec:experiment}.

\subsection{Contributions}
{\it $\bullet$ RNK converges to RNTK uniformly.} Though \cite{tirer2022kernel} showed the RNK point-wisely converge to the RNTK during the training process, it is 
insufficient for showing that  the generalization error of residual networks can be well approximated by that of kernel regression with respect to RNTK. Moreover, their claims do not hold for the ReLU activation.
In this paper, we first show that the RNK converges to the RNTK uniformly and that the dynamic of training the residual network converges to that of the RNTK regression uniformly. Thus, the generalization performance of the residual networks can be approximated well by that of the RNTK regression.

{\it $\bullet$ Generalization performance of residual networks.}
With the assumption that the regression function $f_{\star}\in\mathcal{H}$, the reproducing kernel Hilbert space (RKHS) associated to the RNTK $r$ defined on $\mathbb{S}^{d-1}$, 
we prove that training a residual network with a properly early stopping strategy can produce a residual network achieving the minimax-optimal rate $n^{-d/(2d-1)}$, meaning that the early stopped residual network can generalize well. On the other hand, we can show that if one trains a residual network till it overfits all training data, the resulting network generalizes poorly. 

\subsection{Related works}

There are two lines of works for the convergence of the NNK to the NTK. 
One is assuming that the activation function is a twice differentiable function \cite{jacot2018neural,lee2019wide,tirer2022kernel}. After showing the convergence of the NNK to the NTK, \cite{lee2019wide} proved the convergence of the neural network function to the kernel regression function with the NTK as the width $m\to \infty$. Following the idea of \cite{lee2019wide}, \cite{tirer2022kernel} extend the strategy of \cite{lee2019wide} from fully-connected neural networks to residual networks. However, they only showed the pointwise convergence of NNK and RNK to the corresponding kernel regression functions and the ReLU function does not satisfy their assumption. 
Another one is focusing on the ReLU function. They showed the ``twice differentiability'' of the ReLU function with high probability over random initialization as the width is large enough (see \cite{du2019gradient,allen2019convergence,lai2023generalization,li2023Statistical} in detail). 
\cite{du2019gradient} and \cite{allen2019convergence} proved the pointwise convergence for two-layer fully-connected networks and multilayer fully-connected networks, respectively. Subsequently, \cite{lai2023generalization} and \cite{li2023Statistical} further showed the uniform convergence for two-layer fully-connected networks and multilayer fully-connected networks, respectively. To the best of our knowledge, the uniform convergence of RNK is not solved in the previous papers.

There are a few papers analyzing the generalization error of residual networks through statistical decision theory with different nonparametric regression frameworks. For example, assuming that the data is noiseless and the regression function belongs to the flow-induced function spaces defined in \cite{ma2022barron}, \cite{ma2019generalization} proved the minimum-norm interpolated solution of residual networks can achieve the optimal rate. \cite{ma2020rademacher} showed that there exists a regularized residual network is nearly optimal on the noisy data if the regression function belongs to the Barron space defined in \cite{ma2018priori}. 
Although these empirical risk minimizers are optimal in their corresponding nonparametric regression frameworks, they are hard to apply in practice since the corresponding optimization problems are highly non-linear and non-convex. Thus, these static non-parametric explanations are far from a satisfactory theory. A more practicable theory, such as the generalization theory of residual networks trained by gradient descent, is needed.

\subsection{Preliminaries}
 Let $f_{*}$ be a continuous function defined on a compact subset $\mathcal{X} \subseteq \mathbb{S}^{d-1}$, the $d-1$ dimensional sphere satisfying $\mathbb{S}^{d-1}:=\cl{\bm{x}\in\mathbb R^d:\norm{\bm{x}}_2=1}$. Let $\mu_{\mathcal{X}}$ be a uniform measure supported on $\mathcal{X}$. Suppose that we have observed $n$ i.i.d. samples $\mathcal{D}_n=\cl{\pt{\bm{x}_{i},y_{i}}, i \in [n]}$ sampling from the model:
\begin{equation}\label{equation:true_model}
    y_i=f_{*}(\boldsymbol{x}_i)+\varepsilon_{i}, \quad i=1,\dots,n,
\end{equation}
where $\boldsymbol{x}_{i}$'s are sampled from  $\mu_{\mathcal{X}}$, $\varepsilon_{i} \sim \mathcal{N}(0,\sigma^{2})$ for some fixed $\sigma>0$ and $[n]$ denotes the index set $\{1,2,...,n\}$.
We collect $n$ i.i.d. samples into matrix $\bm X:=(\bm x_1,\dots,\bm x_n)^T\in\mb R^{n\times d}$ and vector $\bm y:=(y_1,\dots,y_n)^T\in\mb R^n$.
We are interested in finding $\hat{f}_{n}$  based on these $n$ samples, which can minimize the excess risk, i.e., the difference between $\mathcal{L}(\hat{f}_{n})=\bm{E}_{(\bm{x},y)}\mbk{\big(\hat{f}_{n}(\bm{x})-y\big)^2}$ and $\mathcal{L}(f_{*})=\bm{E}_{(\bm{x},y)}\left[(f_{*}(\bm{x})-y)^2\right]$.
One can easily verify the following formula about the excess risk:
\begin{equation}\label{eq:mathcal{E}}
\mathcal{E}(\hat{f}_{n})=\mathcal{L}(\hat{f}_{n})-\mathcal{L}(f_{*})=\int_{\mathcal{X}}\pt{\hat{f}_{n}(\bm{x})-f_{*}(\bm{x})}^{2} \mr{d} \mu_{\mathcal{X}}(\bm{x}).
\end{equation}
It is clear that the excess risk is an equivalent evaluation of the generalization performance of $\hat{f}_{n}$. 

\paragraph*{Notation} For two sequence $a_n, b_n,~n\geq 1$ of non-negative numbers,
we write $a_n = O(b_n)$ (or $a_n = \Omega(b_n)$) if there exists absolute constant $C > 0$ such that
$a_n \leq C b_n$ (or $a_n \geq C b_n$).
We also denote $a_n = \Theta(b_n)$ or $a_n \asymp b_n$ if $a_n = O(b_n)$ and $a_n = \Omega(b_n)$ both hold.
We use $\text{poly}(x,y)$ to represent a polynomial of $x,y$ whose coefficients are absolute constants. The ReLU function is denoted as $\sigma(x)=\max(x,0)$.

%% file: s2_NTK.tex
Following the formation of \cite{huang2020deep, belfer2021spectral, tirer2022kernel}, we define an $L$-layer fully connected residual network $f^{m}(\bm{x};\bm{\theta})$ with the width $m$ and the parameters $\bm{\theta}$ in a recursive manner 
\begin{align}
\begin{split}\label{eq:resnet_fuc}
    f^{m}(\bm{x};\bm{\theta}) &=  \bm{v}^T\bm{\alpha}^{(L)}\\
    \bm{\alpha}^{(l)} &= \bm{\alpha}^{(l-1)} + a \sqrt{\frac{1}{m}}\bm{V}^{(l)}\sigma\mpt{\sqrt{\frac{2}{m}} \bm{W}^{(l)}\bm{\alpha}^{(l-1)} }\\
    \bm{\alpha}^{(0)} &=  \sqrt{\frac{1}{m}}\bm{A}\bm{x}
\end{split}
\end{align}
for $l\in[L]$ with parameters $\bm{A}\in\mathbb{R}^{m\times d}$, $ \bm{W}^{(l)}, \bm{V}^{(l)} \in \mathbb{R}^{m\times m}$ and $\bm{v}\in \mathbb{R}^{m}$. All the weights are initialized by the standard normal distribution, i.e., $\bm{v}_i$, $ \bm{V}^{(l)}_{i.j}$, $\bm{W}^{(l)}_{i.j}$, $\bm{A}_{i,k}\iid  \mathcal{N}(0,1)$ for $i,j\in[m], k\in[d], l\in[L]$. The hyper-parameter $a$ is a constant satisfying $0<a<1$ (normally $a=L^{-\gamma}$ with $0.5<\gamma \leq 1$ \cite{huang2020deep,belfer2021spectral}). 

Adopting the derivation in \cite{huang2020deep}, we assume both $\bm{A}$ and $\bm{v}$ are fixed at their initial values and $\bm{W}^{(l)}, \bm{V}^{(l)}$ are learned. Thus, $\bm{\theta}=\text{vec}\mpt{\big\{\bm{W}^{(l)}, \bm{V}^{(l)}\big\}_{l=1}^{L}}$ is the training parameters and the length of $\bm{\theta}$ is  $2Lm^2$. Given $n$ samples $\{(\bm{x}_i,y_i), i\in[n]\}$ from \eqref{equation:true_model}, the network is trained by gradient descent with respect to the empirical loss
\begin{align}
    \hat{\mathcal{L}}_{n}(f^{m}) = \frac{1}{2n}\sum_{i=1}\big(y_i-f^{m}(\bm{x}_i;\bm{\theta})\big)^2.
\end{align}

It is well known that the training of neural networks is a highly non-linear problem. The NTK, a time-invariant kernel proposed by Jacot \cite{jacot2018neural}, can be used to analyze the training process of neural networks as the width $m\rightarrow\infty$. The RNTK, denoted by $r$, is first given in \cite{huang2020deep}. Then \cite{belfer2021spectral} showed that the RNTK $r$ on $\mathbb{S}^{d-1}$ is the inner-product kernel (Theorem 4.1) and simplified  the expression of RNTK on $\mathbb{S}^{d-1}$  as follows (Corollary B.2)
\begin{align}\label{eq:resnet}
    r(\bm{x},\bm{x}') = C\sum_{l=1}^L B_{l+1}(\bm{x},\bm{x}')\bk{(1+a^2)^{l-1}\kappa_1\mpt{\tfrac{K_{l-1}(\bm{x},\bm{x}')}{(1+a^2)^{l-1}}}+K_{l-1}(\bm{x},\bm{x}')\kappa_0\mpt{\tfrac{K_{l-1}(\bm{x},\bm{x}')}{(1+a^2)^{l-1}}}},
\end{align}
where $C=\frac{1}{2L(1+a^2)^{L-1}}$, $K_{0}(\bm{x},\bm{x}') =\bm{x}^T\bm{x}'$, $B_{L+1}(\bm{x},\bm{x}')=1$ and
\begin{gather*}
    \kappa_0(u) = \frac{1}{\pi}(\pi-\arccos u), \quad \kappa_1(u) = \frac{1}{\pi}\left(u(\pi-\arccos u)+\sqrt{1-u^2}\right)\\
    K_{l}(\bm{x},\bm{x}') =K_{l-1}(\bm{x},\bm{x}') + a^2  (1-a^2)^{l-1}\kappa_1\mpt{\frac{K_{l-1}(\bm{x},\bm{x}')}{(1+a^2)^{l-1}}}, \quad l=1,\dots,L-1\\
    B_l(\bm{x},\bm{x}') = B_{l+1}(\bm{x},\bm{x}')\left[1+ a^2\kappa_0\mpt{\frac{K_{l-1}(\bm{x},\bm{x}')}{(1+a^2)^{l-1}}}\right], \quad l=L,\dots,2.
\end{gather*}
\cite{belfer2021spectral} emphasized that for $L=1$, the RNTK $r$ is equal to the NTK of fully-connected neural networks. Thus, we only consider residual networks with $L\geq 2$ in this context.

In this paper, we consider $r(\bm{x},\bm{x}')$ under the uniform measure on $\mathbb{S}^{d-1}$, which admits the following Mercer decomposition:
\begin{align*}
    r(\bm{x},\bm{x}') = a^2\sum_{k=0}^{\infty} \mu_k \sum_{h=1}^{N(d,k)} Y_{k,h}(\bm{x}) Y_{k,h}(\bm{x}'),
\end{align*}
where $N(d,k)$ denotes the number of spherical harmonics of frequency $k$ and $\{Y_{k,h}\}$ are the spherical harmonics on $\mathbb{S}^{d-1}$. \cite{belfer2021spectral} showed the decay rate of $\mu_k$ is $k^{-d}$ for any fixed $L\geq 2$. To better investigate the generalization performance of kernel regression, we rewrite $r(\bm{x},\bm{x}')$ as the following Mercer decomposition:
\begin{align*}
    r(\bm{x},\bm{x}') = \sum_{j=1}^{\infty} \lambda_j \phi_j(\bm{x})\phi_j(\bm{x}'),
\end{align*}
where $\{\lambda_j\}_{j=1}^{\infty}$ and $\{\phi_j\}_{j=1}^{\infty}$ are the decreasing eigenvalues sequence  and  corresponding eigenfunctions sequence of $r(\bm{x},\bm{x}')$ respectively. The decay rate of $\lambda_j$ is more commonly used in analyzing the generalization ability of kernel regression. 
With the decay rate of $\mu_k$, the decay rate of $\lambda_j$ can be derived and guarantees the positive definiteness of the kernel $r$.
\begin{lemma}\label{lem:decay_rate}
Let $\lambda_j$ be the eigenvalues of RNTK $r$ for any fixed $L\geq 2$. Then we have $\lambda_j\asymp j^{-\frac{d}{d-1}}$.
\end{lemma}

\begin{corollary}\label{lem:positive_definite}
$r(\bm{x},\bm{x'})$ is positive definite under the uniform measure on $\mathbb{S}^{d-1}$.
\end{corollary}

The proof of Lemma \ref{lem:decay_rate} is presented in Supplementary Material.

%% file: s3_uniform_convergence.tex
One can consider the empirical loss $\hat{\mathcal{L}}_{n}$ as a function defined on the parameter space, which induces a gradient flow given by
\begin{equation}\label{nn:theta:flow}
\begin{aligned}
      \dot{\bm{\theta}}(t) &=\frac{\partial}{\partial t}\bm{\theta}(t)=-\nabla_{\bm{\theta}}\hat{\mathcal{L}}_{n}(f_{t}^{m})= - \frac{1}{n}\nabla_{\bm{\theta}} f_{t}^{m}(\bm{X}) (f_{t}^{m}(\bm{X})-\bm{y})
\end{aligned}
\end{equation}
where we emphasize that $\nabla_{\bm{\theta}} f_{t}^{m}(\bm{X}) $ is a $2Lm^2\times n$ matrix. When the loss function $\hat{\mathcal{L}}_{n}$ is viewed as a function defined on $\mathcal{F}^{m}$, the space consisting of all residual networks $f_{t}^{m}$, it induces a gradient flow in $\mathcal{F}^{m}$ given by
\begin{equation}\label{nn:f:flow}
\begin{aligned} 
\dot{f}_{t}^{m}(\bm{x}) &=\frac{\partial}{\partial t}f_{t}^{m}(\bm{x})=\nabla_{\bm{\theta}} f_{t}^{m}(\bm{x})^T\dot{\bm{\theta}}(t)= -\frac{1}{n} r_{t}^{m}(\bm{x},\bm{X}) (f_{t}^{m}(\bm{X})-\bm{y}),
\end{aligned}
\end{equation}
where $\nabla_{\bm{\theta}} f_{t}^{m}(\bm{x})$ is a $2Lm^2\times 1$ vector, $r_{t}^{m}(\bm{x},\bm{X}) =\nabla_{\bm{\theta}} f_{t}^{m}(\bm{x})^{T} \nabla_{\bm{\theta}} f_{t}^{m}(\bm{X})$ is a $1\times n$ vector and $r_{t}^{m}$ is a time-varying kernel function
\begin{align*}
\label{eq:vanilla_RNK}
r_{t}^{m}(\bm{x},\bm{x}')=\nabla_{\bm{\theta}}f_{t}^{m}(\bm{x})^{T} \nabla_{\bm{\theta}}f_{t}^{m}(\bm{x}').
\end{align*}

In order to prevent any potential confusion with the RNTK denoted by $r$, we will refer to the time-varying kernel $r_{t}^{m}$ as the RNK in this context.

The gradient flow equations \eqref{nn:theta:flow} and \eqref{nn:f:flow} clearly indicate that the training process of residual networks is influenced by the random initialization of the pararmeters. To maintain focus and avoid unnecessary divergence, we adopt a commonly used initialization method from the existing literature \cite{hu2019simple, chizat2019lazy,lai2023generalization,li2023Statistical}, which ensures that $f_{0}^{m}(\bm{x})$ is initialized to zero. The detail is shown in Appendix. 

It is well-known that the explicit solution of the highly non-linear equations \eqref{nn:theta:flow} and  \eqref{nn:f:flow} is hard to find out. But one can follow the idea of the NTK approach, using the NTK regression solutions to characterize the asymptotic behavior of the exact solution of these equations. For example, the time-independent kernel RNTK
offered us a simplified version of the equation  \eqref{nn:f:flow}:
\begin{equation}\label{ntk:f:flow}
    \begin{aligned}
    \dot{f}^{NTK}_{t}(\bm{x})=\frac{\partial}{\partial t}f^{NTK}_{t}(\bm{x})=-\frac{1}{n}r(\bm{x},\bm{X})(f^{NTK}_{t}(\bm{X})-\bm{y})
    \end{aligned}
\end{equation}
where  $r(\bm{x},\bm{X}) = (r(\bm{x},\bm{x}_1),\dots,r(\bm{x},\bm{x}_n))\in \mathbb{R}^{1\times n}$.
This equation is defined on the space $\mathcal{H}$, the RKHS associated to the kernel $r$. 
The equation \eqref{ntk:f:flow} is called the gradient flow associated to the kernel regression with respect to the kernel $r$. 
Similar to the special initialization of the residual network function, we assume that the initial function $f^{NTK}_{0}(\bm{x})=0$. Then the equation \eqref{ntk:f:flow}  can be solved explicitly:
\begin{equation}\label{ntk:solution}
    f_t^{NTK}(\bm{x})=r(\bm{x},\bm{X})r(\bm{X},\bm{X})^{-1}(\bm{I}-\me^{-\frac{1}{n}r(\bm{X},\bm{X})t})\bm{y}
\end{equation}
where $r(\bm X,\bm X):=\{r(\bm x_i,\bm x_j)\}_{i,j=1}^n\in\mb R^{n\times n}$.

\cite{tirer2022kernel} showed the RNK point-wisely converges to the RNTK during the training process, i.e., for given $\bm{x}, \bm{x'}\in \mathcal{X}$, $\sup_{t\geq 0} |r^{m}_t(\bm{x},\bm{x}')-r(\bm{x},\bm{x}')|\to 0$ with high probability as the width $m\to \infty$. However, it is 
insufficient for showing that the generalization error of residual networks can be well approximated by that of kernel regression with respect to RNTK. 
Moreover,  \cite{tirer2022kernel} assumed the activation function satisfies that for $\forall x,x'\in \mathbb{R}$, $|\sigma'(x)-\sigma'(x')| \leq C |x-x'|$ for some constant $C>0$, which is not applicable to the ReLU function.


Let us denote by $\lambda_0 = \lambda_{\min}\left( r(\bm{X},\bm{X}) \right)$ the minimal eigenvalue of the kernel matrix $r$. Corollary \ref{lem:positive_definite} has shown that $r$ is positive definite and thus $\lambda_0 > 0$ almost surely. One of our main technical contributions is that the convergence of the RNK $r^{m}_t$ to the RNTK $r$ is uniform with respect to all $t\geq0$ and all $\bm{x}\in\mathcal{X}$. Thus, the excess risk $\mathcal{E}(f_{t}^{m})$ of the wide residual network $f_{t}^{m}$ could be well approximated by the excess risk $\mathcal{E}(f_{t}^{NTK})$ of the RNTK regression function $f_{t}^{NTK}$. 

\begin{theorem}\label{thm:risk:approx}
There exists a polynomial $\operatorname{poly}(\cdot):\mb R^5\to\mb R$, such that for any given  training data $\{(\bm{x}_{i},y_{i}),i\in[n]\}$,  any $\epsilon>0$ and any $\delta\in(0,1)$, when the width $m\geq \operatorname{poly}(n,\lambda_0^{-1},\|\bm{y}\|_2,\log(1/\delta), 1/\epsilon)$, we have 
    \begin{equation*}
        \sup_{t\geq 0}|\mathcal{E}(f_{t}^{m})-\mathcal{E}(f_{t}^{NTK})|\leq \epsilon 
    \end{equation*}
    holds with probability at least $1-\delta$ with respect to the random initialization.
\end{theorem}

The key to the proof of Theorem \ref{thm:risk:approx} is to show the uniform convergence of the RNK $r_{t}^{m}$, which is given by the following proposition:
\begin{proposition}\label{prop:kernel:kernel}
There exists a polynomial $\operatorname{poly}(\cdot):\mb R^4\to\mb R$, such that for any given  training data $\{(\bm{x}_{i},y_{i}),i\in[n]\}$ and any $\delta\in(0,1)$, when the width $m\geq \operatorname{poly}(n,\lambda_0^{-1},\|\bm{y}\|_2,\log(1/\delta))$,  we have
\begin{align}
    \sup_{t\geq 0} \sup_{\bm{x},\bm{x}'\in \mathcal{X}}|r_t^{m}(\bm{x},\bm{x}')-r(\bm{x},\bm{x}')|\leq O(m^{-\frac{1}{12}}\sqrt{\log (m)}),
\end{align}
with probability at least $1-\delta$.
\end{proposition}
By  applying the proof strategy in \cite[Proposition 3.2, Theorem 3.1]{lai2023generalization}, we can utilize the  Proposition 3.2 in the current paper to finish the proof of Theorem \ref{thm:risk:approx}.

Proposition \ref{prop:kernel:kernel} shows the uniform convergence of the kernel for any $\bm{x},\bm{x}'\in \mathcal{X}\subseteq \mathbb{S}^{d-1}$. 
The proof of Proposition \ref{prop:kernel:kernel} is divided into three parts: $i)$ we show that the RNK $r_t^{m}$ point-wisely converges to the RNTK $r$,
$ii)$ we prove the H\"{o}lder continuity of $r_t^{m}$ and the H\"{o}lder continuity of $r$;
$iii)$ we use the $\epsilon$-net argument to present the uniform convergence of $r_t^{m}$ to $r$. Proposition \ref{prop:kernel:kernel} extends the ``fully-connected network version‘’ that appears in \cite{li2023Statistical} to the considered residual network. Yet, since the structure of the residual network is much more complex than that of the fully-connected network, the proof of Proposition \ref{prop:kernel:kernel} is more challenging. The proof of Proposition \ref{prop:kernel:kernel} is presented in Supplementary Material.

%% file: s4_generalization.tex

To facilitate a comprehensive analysis of the generalization performance of a residual network, it is essential to define the class of functions to which $f_{\star}$ belongs. In this study, we introduce the following assumption:

\begin{assumption}\label{assump:f_star}
    The regression function $f_{*}\in \mathcal{H}$ and $\| f_{*}\|_{\mathcal{H}}\leq R$ for some constant $R$, where $\mathcal{H}$ is the RKHS associated to the kernel $r$.
\end{assumption}

Theorem \ref{thm:risk:approx} shows that $\mathcal{E}(f_{t}^{m})$ is well approximated by $\mathcal{E}(f_{t}^{NTK})$. Thus we can focus on studying the generalization ability of the RNTK regression function. Assumption \ref{assump:f_star} is actually a usual assumption appeared in the kernel regression literature (see e.g., \cite{caponnetto2007optimal, yao2007early, raskutti2014early, blanchard2018optimal, lin2020optimal}).

\subsection{Wide residual networks with early stopping achieve the minimax rate}

The early stopping strategy is a widely employed technique in the training of various models, including kernel regression and neural networks, among others. Extensive research studies have provided solid theoretical foundations for the efficacy of early stopping \cite{yao2007early, raskutti2014early, blanchard2018optimal, lin2020optimal}, where the determination of the optimal stopping time relies on the decay rate of eigenvalues associated with the kernel.
It is worth noting that Corollary \ref{lem:decay_rate} furnishes us with the decay rate of eigenvalues for the kernel $r$, while Theorem \ref{thm:risk:approx} establishes the assurance that the excess risk of the RNTK regression function $f_{t}^{NTK}$ provides a reliable approximation of the excess risk of the residual network $f_{t}^{m}$. As a result, we can derive the following Theorem \ref{thm:early_stopping}.

\begin{theorem}[Early-stopped residual networks can generalize]\label{thm:early_stopping}
Suppose Assumption \ref{assump:f_star} holds. For any given $\delta\in(0,1)$, if one trains a residual network with width $m$ that is sufficiently large and stops the gradient flow at time $t_{*}\propto n^{d/(2d-1)}$, then for sufficiently large $n$, there exists a constant $C$ independent of $\delta$ and $n$, such that   
    \begin{equation}\label{eq:mathcal{E} nn}
\mathcal{E}(f_{t_{*}})\leq Cn^{-\frac{d}{2d-1}}\log^{2}\frac{6}{\delta}
    \end{equation}
    holds with probability at least $1-\delta$.
\end{theorem}

\cite{blanchard2018optimal} established the following minimax rate of regression over the RKHS $\mathcal{H}$ associated to $r$:
    \begin{equation}    \inf_{\hat{f}_{n}}\sup_{f_{*}\in\mathcal{H},\| f_{*}\|_{\mathcal{H}\leq R}}   \bm{E}\mathcal{E}(\hat{f}_{n})=\Omega(n^{-\frac{d}{2d-1}}).
    \end{equation}
Thus, we have proved that training a wide residual network with the early stopping strategy achieves the optimal rate. The proof of Theorem \ref{thm:early_stopping} can be found in Supplementary Material.

\subsection{Overfitted residual networks generalize poorly}

In this subsection, we are more interested in the generalization performance of $f_{t}^{m}(x)$ for sufficiently large $t$ such that $f_{t}^{m}(x)$ can (nearly) fit the given data. As $t\to \infty$, Equation \eqref{ntk:solution} is considered as the kernel interpolation. Theorem 3.1 in \cite{li2023kernel} showed that the kernel interpolation generalizes poorly. The following theorem, which is a consequence of Theorem \ref{thm:risk:approx} in the current paper and Theorem 3.1 in \cite{li2023kernel}, shows overfitted residual networks generalize poorly

\begin{theorem}[Overfitted residual networks generalize poorly]\label{thm:overfitting}
For any $\epsilon > 0$ and $\delta \in (0,1)$,
  there is some constant $c> 0$ such that
  when $n$ and $m$ are sufficiently large, with the probability at least $1-\delta$, we have
  \begin{align*}
    \bm{E} \left[\liminf_{t\to\infty} \mathcal{E}(f_{t}^{m}) \;\Big|\; \boldsymbol X \right]  \geq c n^{-\epsilon}.
  \end{align*}
\end{theorem}
Though this theorem can not imply that the kernel interpolation is inconsistent, it actually shows that overfitted residual networks generalize poorly, which contradicts the ``benign overfitting phenomenon''. 
In the next section, we will illustrate  several experiments on residual networks to reconcile this contradiction and show that the occurrence of ``benign overfitting phenomenon'' depends on the signal strength of the data.

%% file: s5_experiment.tex
In Section \ref{sec:generalization}, we have shown that the generalization error of a residual network depends on the stopping time.  $i)$ the residual network with the proper stopping time can achieve the minimax rate; $ii)$ the residual network trained till the loss is near zero can not generalize well. However, the second result seems contradict to the reported ``benign overfitting phenomenon'' where overfitted neural networks do generalize well in many datasets. Thus, for residual networks, we need to find an explanation that can reconcile the conflict between the second result and the widely observed ``benign overfitting phenomenon''.



To reconcile the same conflict for fully-connected neural networks, \cite{lai2023generalization} has justified an insightful hypothesis. They emphasized that a subtle difference between the classification problem and the regression problem might be ignored in the reported experiments (Figure 2 in \cite{lai2023generalization}). For the classification problem, they denoted $i)$ $t_{\text{opt}}$ as the optimal early stopping time, $ii)$ $t_{\text{loss}}$ as the time when the value of the loss function nears zero and $iii)$ $t_{\text{label}}$ as the time when the value of the label error rate nears zero. They notice that  
most of the reported experiments in ``benign overfitting phenomenon'' utilize the stopping time $t_{\text{label}}$ and claim that the resulting neural network can overfit the data and generalize well \cite{nakkiran2021deep,zhang2016understanding}. Thus \cite{lai2023generalization} justified the following hypothesis for the fully-connected network: $i)$ if the signal strength is strong, then $t_{\text{label}}$ nears $t_{\text{opt}}$  and $t_{\text{label}}$ is much earlier than $t_{\text{loss}}$ ; $ii)$ if the signal strength is weak, then $t_{\text{label}}$ is away from $t_{\text{opt}}$ and nears $t_{\text{loss}}$, where we can consider $t_{\text{loss}}=\infty$.  Adopting the same idea, we justify this hypothesis through various experiments for residual networks.

\begin{figure}[htbp]
\centering
\begin{minipage}[t]{0.4\linewidth}
      \centering
      \includegraphics[width=\textwidth]{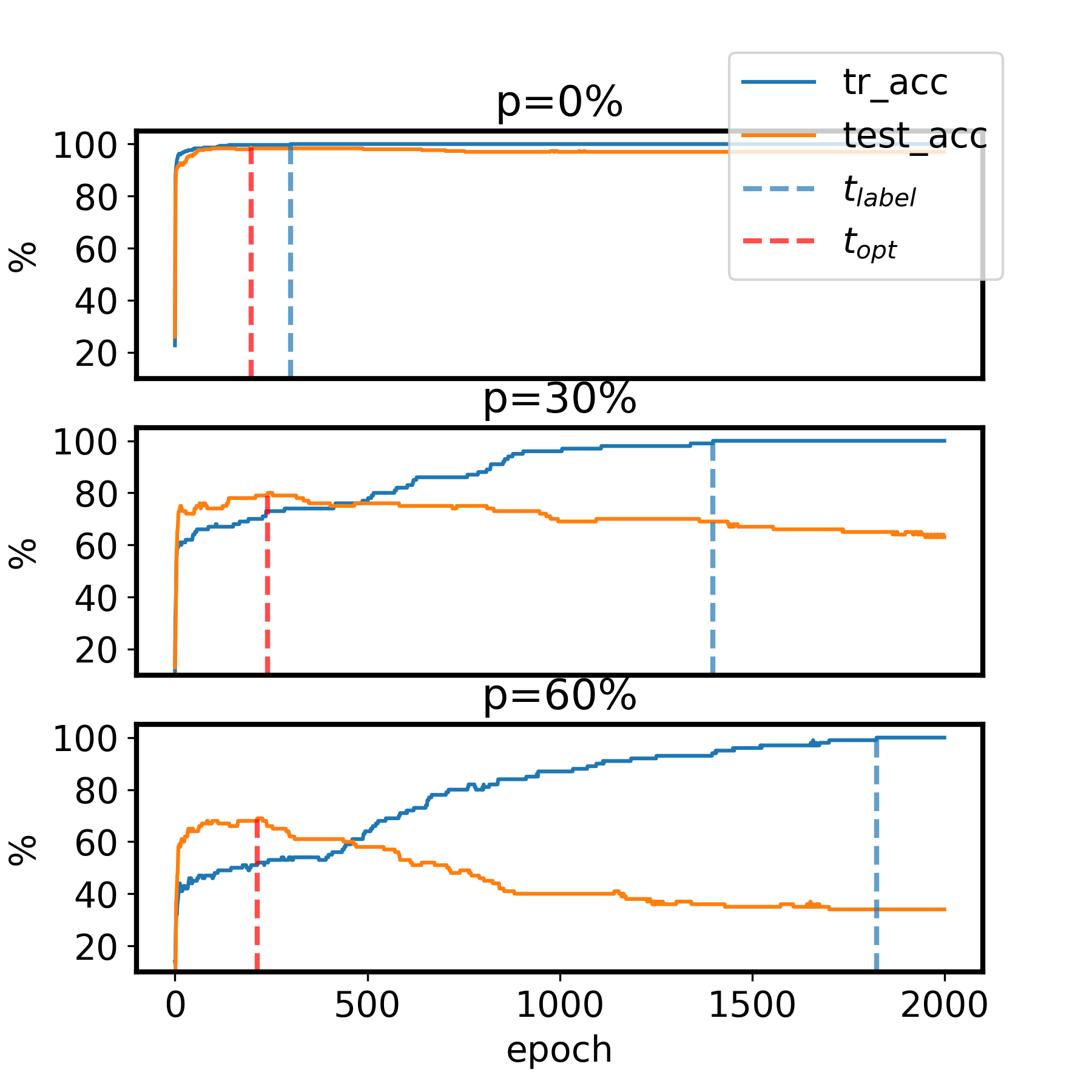}
  \end{minipage}
  \begin{minipage}[t]{0.4\linewidth}
      \centering
      \includegraphics[width=\textwidth]{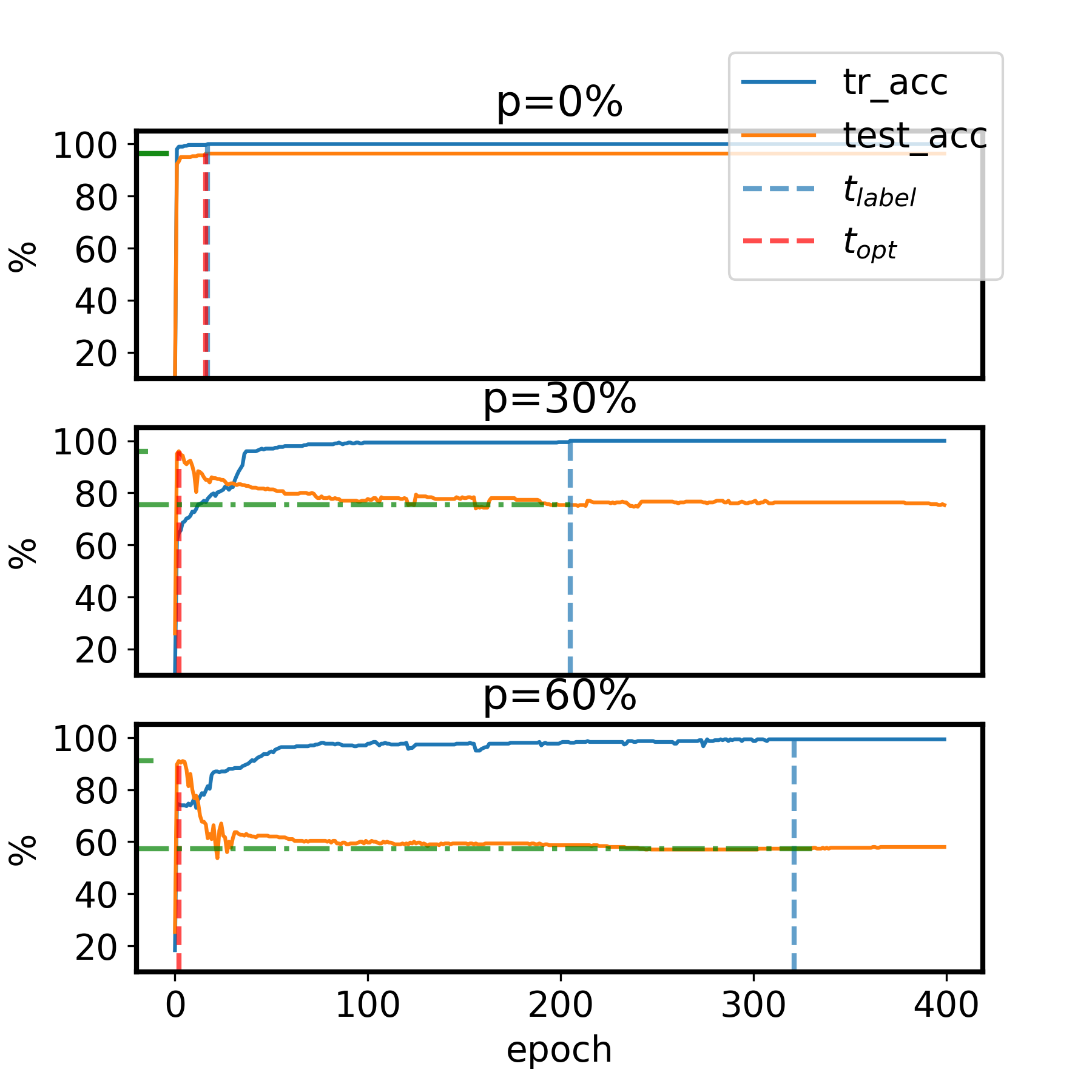}
  \end{minipage}\\
  \begin{minipage}[t]{0.4\linewidth}
      \centering
      \includegraphics[width=\textwidth]{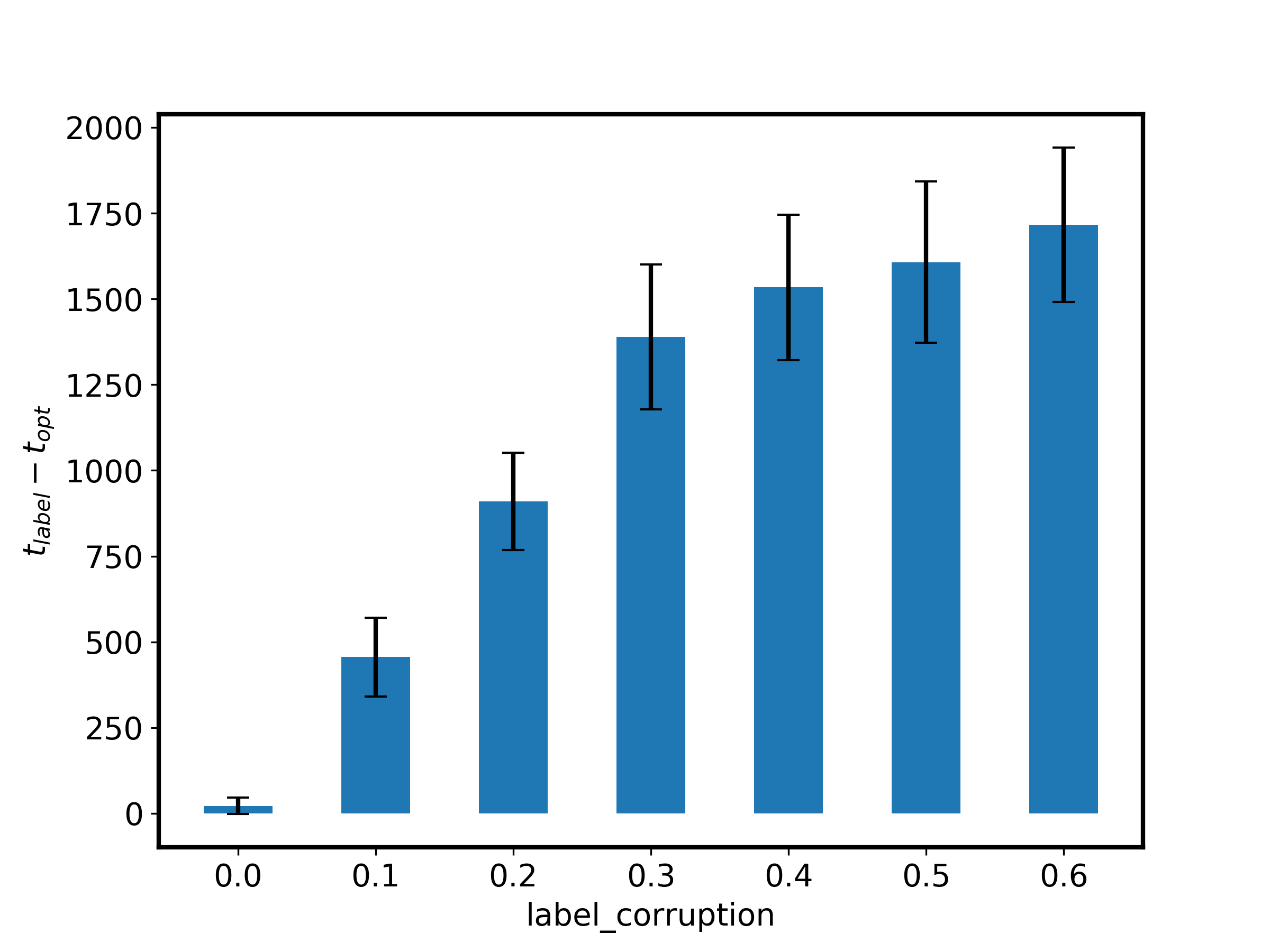}
      \centerline{(a) the results of the square loss}
  \end{minipage}
  \begin{minipage}[t]{0.4\linewidth}
      \centering
      \includegraphics[width=\textwidth]{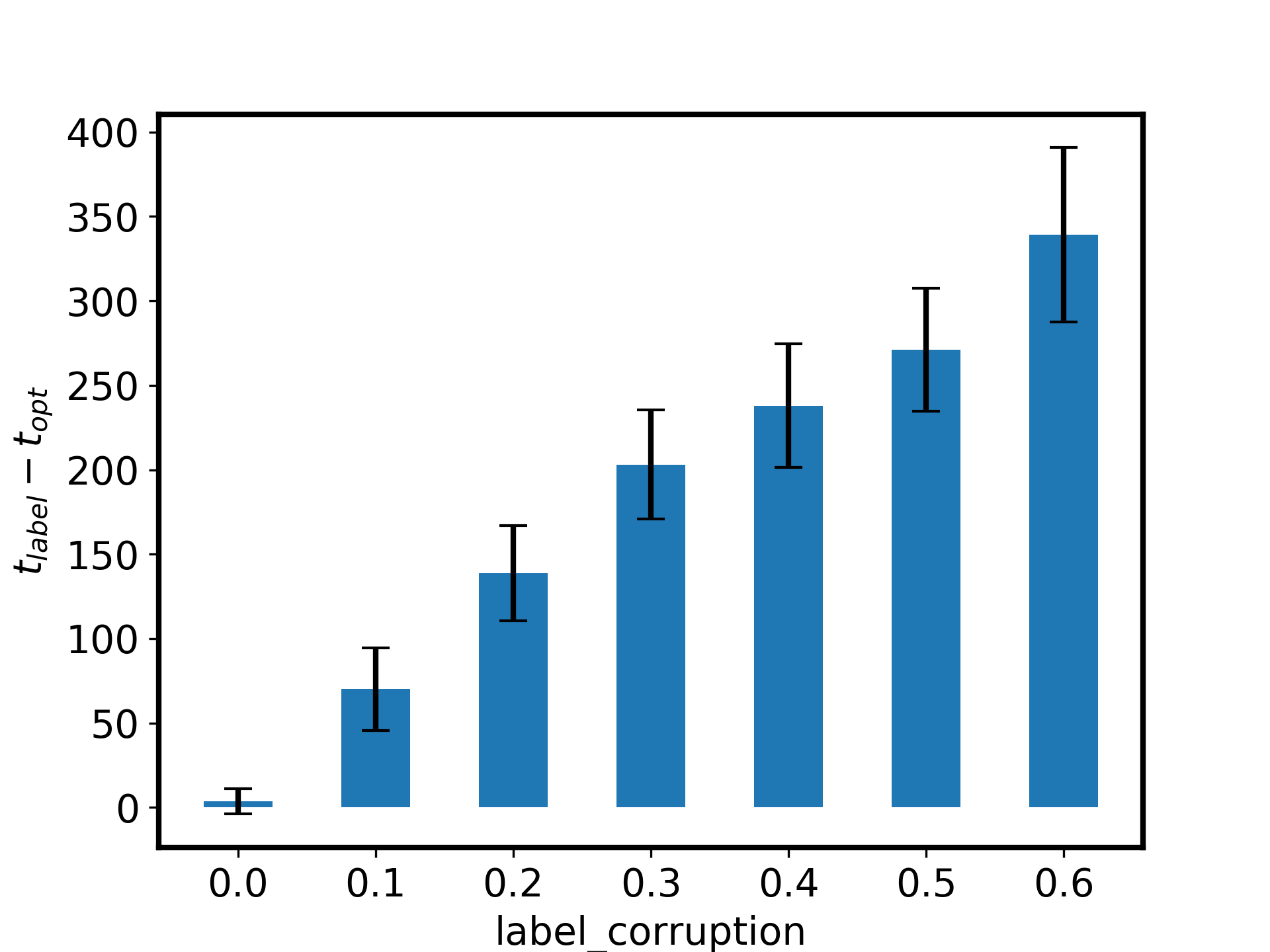}
      \centerline{(b) the results of the cross-entropy loss}
  \end{minipage}
  \caption{Synthetic Data: the gap between $t_{\text{label}}$ and $t_{\text{opt}}$ is increasing when the label corruption ratio $p$ is increasing. The lower figures are the error bars associated with 10 replicate experiments of two different losses and show the mean and the standard derivation of the gap between $t_{\text{label}}$ and $t_{\text{opt}}$. When $p=0$, the time gap between $t_{\text{label}}$ and $t_{\text{opt}}$ and the gap between the corresponding testing accuracy is extremely small, i.e., we observed the ``benign overfitting''.}
  \label{fig: diff_noise}
\end{figure}

\noindent{\it  $\bullet$ Synthetic Data:} 
Suppose that $\bm{x}_i , 1\leq i \leq 500$ are i.i.d. uniformly sampled from $\mathbb{S}^{2}$ and 
\begin{equation*}
    y_{i}=f_{\star}(\bm x_{i})=\lfloor \bm{x}_{i,(1)}+1\rfloor+2\lfloor \bm{x}_{i,(2)}+1\rfloor+4\lfloor \bm{x}_{i,(3)}+1\rfloor\in \{0,1,\cdots,7\}.
\end{equation*} 
For a given $p\in [0,1]$, we corrupt every label $y_{i}$ of the data with probability $p$ by a uniform random integer from $\{0,1,\cdots,7\}$. For corrupted data with $p\in\{0, 0.1,0.2,0.3,0.4,0.5,0.6\}$, we train a residual fully-connected network (Equation \eqref{eq:resnet_fuc} with $m=1000$, $L=5$, $a=0.5$) with the squared loss and the cross-entropy loss. We collect the testing accuracy based on $10000$ testing data points with label corruption. The results are reported in Figure \ref{fig: diff_noise}.

\vspace{3mm}
{\it \noindent $\bullet$ Real Data:} We perform the experiments on CIFAR-10 with the convolutional residual network. We use the model architecture introduced in \cite[Section 4.2]{he2016deep}. The first layer is $3\times3$ convolutions with 32 filters. Then we use a stack of 6 layers with $3\times3$ convolutions, 2 layers with 32 filters, 2 layers with 64 filters and 2 layers with 128 filters. The network ends
with a global average pooling and a fully-connected layer. There are 8 stacked weighted
layers in total. 

\begin{figure}[htbp]
\begin{minipage}[t]{0.5\linewidth}
      \centering
      \includegraphics[width=\textwidth]{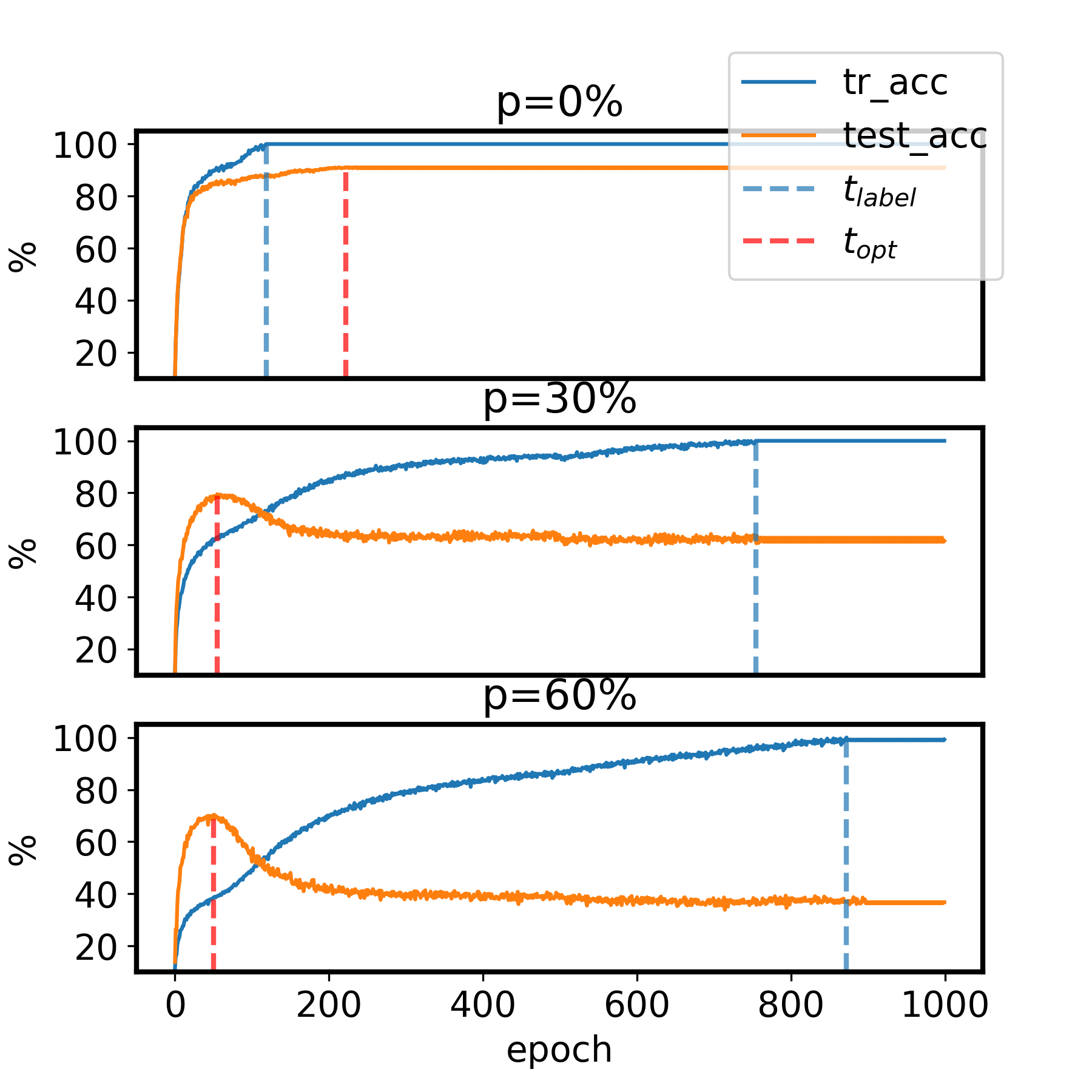}
  \end{minipage}
  \begin{minipage}[t]{0.5\linewidth}
      \centering
      \includegraphics[width=\textwidth]{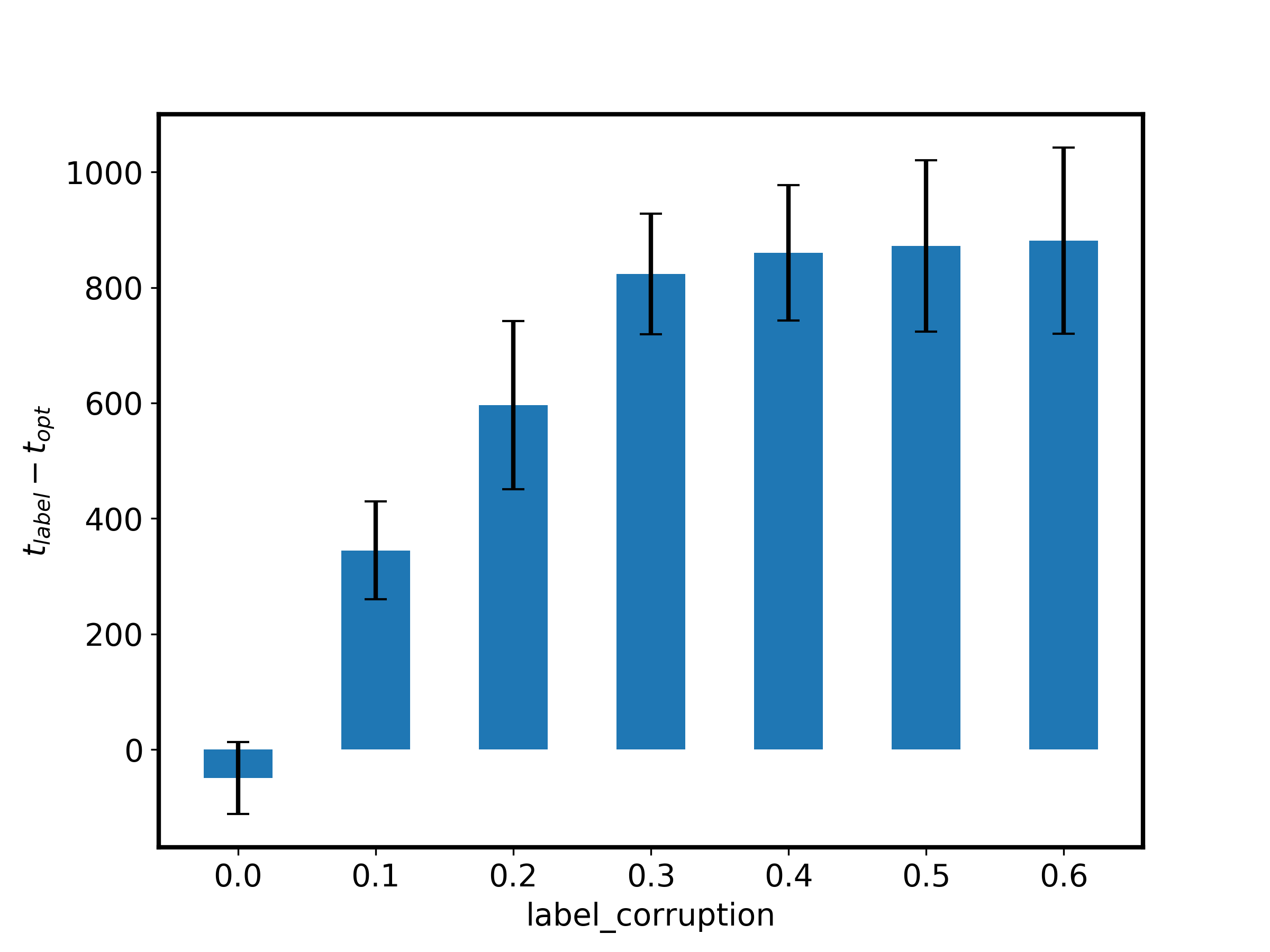}
  \end{minipage}
  \caption{Convolutional residual network on CIFAR-10 (cross-entropy loss): the time gap between $t_{label}$ and $t_{opt}$ is increasing when the label corruption ratio is increasing.}
  \label{fig: cifar10_diff_noise_ce}
\end{figure}

We corrupt the data with $p\in\{0, 0.1,0.2,0.3,0.4,0.5,0.6\}$ and apply the Adam to training Alex with the initial learning rate of 0.001 and the decay factor 0.95 per training epoch. The results are reported in Figure \ref{fig: cifar10_diff_noise_ce}. 


The experimental results presented above provide empirical evidence that supports our hypothesis and resolves the discrepancy between the observed "benign overfitting phenomenon" and our theoretical framework for residual networks. Specifically,  when the signal strength is strong, the "benign overfitting phenomenon" holds, and our theoretical results remain valid. However, when the signal strength is weak, the "benign overfitting" phenomenon no longer holds, and our theoretical findings offer an explanation for the failure of the "benign overfitting phenomenon."

%% file: s6_conclusion.tex


In this study, we demonstrated that the RNK uniformly converges to the RNTK, indicating that kernel regression using the RNTK can effectively approximate the excess risk of residual networks. By analyzing the decay rate of eigenvalues associated with the RNTK, we established two key results: $i)$ stopping the training process of residual networks at a suitable time can lead to a resulting neural network with excess risk achieving minimax optimality, and $ii)$ an overfitted residual network may not generalize well. Additionally, we conducted experiments to address the discrepancy between our theoretical findings and the commonly observed "benign overfitting phenomenon."

Drawing on the approach of \cite{lai2023generalization}, our strategy can be applied to various neural network architectures. Specifically, one can first demonstrate the uniform convergence of neural network kernels (e.g., convolutional neural networks and recurrent neural networks) to the corresponding neural tangent kernels, and then examine the spectral properties of the NTK, including positive definiteness and eigenvalue decay rate. Thus, we anticipate that networks using an early stopping strategy can achieve optimal minimax rates.

%% file: sa_Proof_uniform_convergence.tex
\section{Proof of Proposition 3.2}

\begin{figure}[htbp]
 \centering
 \includegraphics[width=0.6\textwidth]{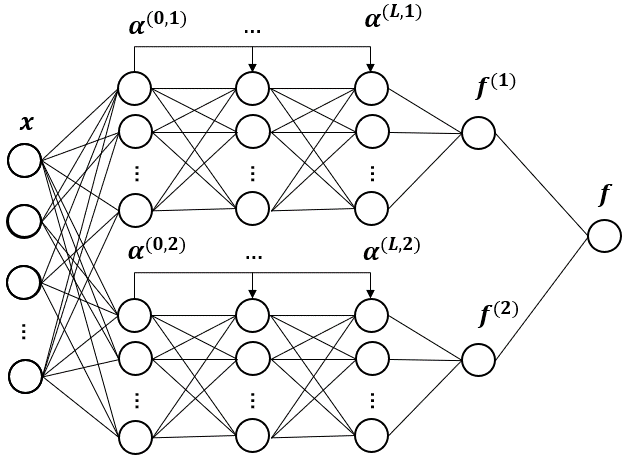}
 \caption{Special initialization}
 \label{fig:special_initialization}
 \end{figure}

We define an $L$-layer fully connected residual network with the following parameterization (Figure \ref{fig:special_initialization})
\begin{align}
\begin{split}\label{eq:resnet_fuc_mirror}
f^{m}\mpt{\bm{x};\bm{\theta}} &= \frac{\sqrt{2}}{2}\bk{f^{(1),m}\mpt{\bm{x};\bm{\theta}^{(1)}}-f^{(2),m}\mpt{\bm{x};\bm{\theta}^{(2)}}};\\
f^{(p),m}\mpt{\bm{x};\bm{\theta}^{(p)}} &= \pt{\bm{v}^{(p)}}^T\bm{\alpha}^{(p,L)}; \\
\bm{\alpha}^{(p,l)} &= \bm{\alpha}^{(p,l-1)} + a \sqrt{\frac{1}{m}}\bm{V}^{(p,l)}\sigma\mpt{\sqrt{\frac{2}{m}} \bm{W}^{(p,l)}\bm{\alpha}^{(p,l-1)}};\\
\bm{\alpha}^{(p,0)}&= \sqrt{\frac{1}{m}}\bm{A}^{(p)}\bm{x},
\end{split}
\end{align}
where $p=1,2$ and $l\in[L]$. 

\paragraph*{Initialization} We initialize the parameters as followed:
\begin{align*}
\begin{gathered}
\text{for}~i,j\in[m],~k\in[d],~l\in[L]:\quad~\bm{A}^{(1)}_{i,k},~ \bm{W}^{(1,l)}_{i,j},~\bm{V}^{(1,l)}_{i,j},~ \bm{v}^{(1)}_{i} \iid \mathcal N(0,1);\\
\bm{A}^{(1)} = \bm{A}^{(2)},\quad~ \bm{W}^{(1,l)} = \bm{W}^{(2,l)},\quad~ \bm{V}^{(1,l)} = \bm{V}^{(2,l)}, \quad~\bm{v}^{(1)} = \bm{v}^{(2)}.
\end{gathered}
\end{align*}
Thus, we have $f^{m}_0(\bm{x};\bm{\theta})=0$ for any $\bm{x}\in \mathbb{S}^{d-1}$. 

\paragraph*{Training} Let us consider the empirical square loss
\begin{align}
 \hat{\mathcal{L}}_{n}(f^{m}) = \frac{1}{2n}\sum_{i=1}^{n}\big(y_i-f^{m}(\bm{x}_i;\bm{\theta})\big)^2.
\end{align}
Denoting by $\bm{\theta}_t$ the parameter at the time $t \geq 0$, we train the network by the gradient flow
\begin{equation}
\begin{aligned}
 \dot{\bm{\theta}}(t) &=-\nabla_{\bm{\theta}}\hat{\mathcal{L}}_{n}(f_{t}^{m})= - \frac{1}{n}\nabla_{\bm{\theta}} f_{t}^{m}(\bm{X}) (f_{t}^{m}(\bm{X})-\bm{y})
\end{aligned}
\end{equation}
where we emphasize that $\nabla_{\bm{\theta}} f_{t}^{m}(\bm{X}) $ is a $2Lm^2\times n$ matrix. Let us denote by $f_t(\bm x) = f(\bm x;\bm{\theta}_t)$ the resulting neural network function.

\paragraph*{Neural network kernel} With the above architecture, we have
\begin{align*}
 r_t^{m}\mpt{\bm{x},\bm{x}'} & = \ag{\nabla_{\bm{\theta}} f_t(\bm{x};\bm{\theta}_t), \nabla_{\bm{\theta}} f_t(\bm{x}';\bm{\theta}_t) }\\
 &= \frac{1}{2}\sum_{p=1}^2 \ag{\nabla_{\bm{\theta}^{(p)}} \mpt{f^{(1),m}_t(\bm{x}) - f^{(2),m}_t(\bm{x})},
 \nabla_{\bm{\theta}^{(p)}} \mpt{f^{(1),m}_t(\bm{x}) - f^{(2),m}_t(\bm{x})}} \\
 &= \frac{1}{2} \sum_{p=1}^2 \ag{\nabla_{\bm{\theta}^{(p)}} f^{(p),m}_t(\bm{x}),\nabla_{\bm{\theta}^{(p)}} f^{(p),m}_t(\bm{x}') } = \frac{1}{2} \pt{r_t^{(1),m}(\bm{x},\bm{x}') + r_t^{(2),m}(\bm{x},\bm{x}')},
\end{align*}
where $r_t^{(p),m}(\bm{x},\bm{x}')$ is the RNK of $f_t^{(p),m}$, which is the vanilla neural network.
Consequently, we have
\begin{align*}
 r_0^{(1),m}(\bm{x},\bm{x}') = r_0^{(2),m}(\bm{x},\bm{x}')= r_0^{m}(\bm{x},\bm{x}').
\end{align*}
Thus, we only need to show the uniform convergence of one of the RNKs $\big\{r_0^{(p),m}\big\}_{p=1,2}$ since another RNK has the same uniform convergence.

\paragraph*{Further notations}
To simplify the notation, except for $f^{(p),m}$ and $r^{(p),m}$, we sometimes omit the index $p$ on the parameters $\bm{W}^{(l)}$, $\bm{V}^{(l)}$, $\bm{A}$, $\bm{v}$ and their derived notations. If there is no additional statement, the conclusions hold for $p=1,2$.

For $l\in\cl{0,1,\cdots,L}$, denote $\bm{\delta}^{(l)}(\bm x)=\nabla_{\bm{\alpha}^{(l)}}f^{(p),m}(\bm x)=\nabla_{\bm{\alpha}^{(l)}}\bm{\alpha}^{(L)}(\bm x)\bm{v}$. It is easy to check that
\begin{align*}
\bm{\delta}^{(l)}(\bm x)=
\begin{cases}
\nabla_{\bm{\alpha}^{(l)}}\bm{\alpha}^{(l+1)}(\bm x)\bm{\delta}^{(l+1)}(\bm x),~& l=0,1,\cdots,L-1;\\
\bm{v},~& l=L,
\end{cases}
\end{align*}
where 
\begin{align*}
\nabla_{\bm{\alpha}^{(l-1)}}{\bm{\alpha}^{(l)}}(\bm x)=\pt{\bm{I}_m + \frac{\sqrt 2 a}{m} \bm{V}^{(l)} \bm{D}^{(l)}(\bm x) \bm{W}^{(l)}}^T\qquad~\text{for}~l\in[L]
\end{align*}
and
\begin{align*}
 \bm{D}^{(l)}(\bm{x}) = \text{diag}\mpt{\sigma'\mpt{ \sqrt{\frac{2}{m}} \bm{W}^{(l)}\bm{\alpha}^{(l-1)}(\bm x)}}\qquad~\text{for}~l\in[L].
\end{align*}

The gradient of $\bm{W}^{(l)}$ and $\bm{V}^{(l)}$ can be presented as followed:
\begin{align}
\begin{split}\label{eq: nabla_W_V_f}
&\nabla_{\bm{W}^{(l)}} f^{(p),m}(\bm{x}) = \frac{\sqrt 2 a}{m} \bm{D}^{(l)}(\bm x) \bm{V}^{(l),T} \bm{\delta}^{(l)}\bm{\alpha}^{(l-1),T}=a\bm{\gamma}^{(l)}(\bm x)\bm{\alpha}^{(l-1),T}(\bm x);\\
&\nabla_{\bm{V}^{(l)}} f^{(p),m}(\bm{x}) = \frac{\sqrt 2 a}{m} \bm{\delta}^{(l)}(\bm x) \bk{\sigma\mpt{\bm{W}^{(l)} \bm{\alpha}^{(l-1)}(\bm x)}}^T =a\bm{\delta}^{(l)}(\bm x)\bm{\eta}^{(l),T}(\bm x),
\end{split}
\end{align}
where 
\begin{align*}
 \bm{\gamma}^{(l)}(\bm x)=\frac{\sqrt 2}{m}\bm{D}^{(l)}(\bm x) \bm{V}^{(l),T} \bm{\delta}^{(l)}(\bm x);\qquad~\bm{\eta}^{(l)}(\bm x) = \frac{\sqrt{2}}{m}\bm{D}^{(l)}(\bm x)\bm{W}^{(l)}\bm{\alpha}^{(l-1)}(\bm x).
\end{align*}
And the RNK can be formulated as 
\begin{align}
\begin{split}\label{eq: RNK_formulate}
 r_t^{(p),m}(\bm{x},\bm{x}') =\sum_{l=1}^{L}\Big(&\ag{ \nabla_{\bm{W}^{(l)}} f^{(p),m}_t(\bm{x}), \nabla_{\bm{W}^{(l)}} f^{(p),m}_t(\bm{x}')} \\
 +& \ag{\nabla_{\bm{V}^{(l)}} f^{(p),m}_t(\bm{x}), \nabla_{\bm{V}^{(l)}} f^{(p),m}_t(\bm{x}')}\Big).
\end{split}
\end{align}

To shorten the notations, we denote
\begin{align*}
\begin{gathered}
 \bm{\delta}^{(l)}_{t,\bm x} = \bm{\delta}^{(l)}_t(\bm x), \quad \bm{\alpha}^{(l)}_{t,\bm x} = \bm{\alpha}^{(l)}_t(\bm x), \quad \bm{D}^{(l)}_{t,\bm x} = \bm{D}^{(l)}_t(\bm x), \quad \bm{\gamma}^{(l)}_{t,\bm x} = \bm{\gamma}^{(l)}_t(\bm x), \quad \bm{\eta}^{(l)}_{t,\bm x} = \bm{\eta}^{(l)}_t(\bm x).\\
 \Delta\bm{\delta}^{(l)}_{\bm x \bm z} :=\bm{\delta}^{(l)}_{t,\bm x} -\bm{\delta}^{(l)}_{0,\bm z}, \quad \Delta\bm{\alpha}^{(l)}_{\bm x \bm z}={\bm{\alpha}^{(l)}_{t,\bm x} -\bm{\alpha}^{(l)}_{0,\bm z}}, \quad \Delta\bm{\gamma}^{(l)}_{\bm x \bm z} :=\bm{\gamma}^{(l)}_{t,\bm x} -\bm{\gamma}^{(l)}_{0,\bm z},\quad \Delta\bm{\eta}^{(l)}_{\bm x \bm z} :=\bm{\eta}^{(l)}_{t,\bm x} -\bm{\eta}^{(l)}_{0,\bm z}
 \end{gathered}
\end{align*}
and
\begin{align*}
\begin{gathered}
\bm{D}^{(l)\prime}_{\bm x \bm z}=\bm{D}^{(l)}_{t,\bm z} - \bm{D}^{(l)}_{0,\bm z}, \quad 
 \bm{g}_{l,\bm x \bm z}'=\sqrt{\frac{2}{m}} \bm{W}^{(l)}_t \bm{\alpha}^{(l-1)}_{t,\bm x}- \sqrt{\frac{2}{m}}\bm{W}^{(l)}_0 \bm{\alpha}^{(l-1)}_{0,\bm z},\\
 \Delta\bm{V}^{(l)}=\bm{V}^{(l)}_t -\bm{V}^{(l)}_0;\quad~ \Delta\bm{W}^{(l)}=\bm{W}^{(l)}_t -\bm{W}^{(l)}_0.
 \end{gathered}
\end{align*}


\subsection{Initialization}

\begin{theorem}[Theorem 4 of \cite{huang2020deep}] \label{thm:initialization of NTK}There exist some positive absolute constants $C_1>0$ and $C_2\geq 1$, such that if $\varepsilon\in\pt{0,1/2}$, $\delta\in\pt{0,1}$ and $m\geq C_1\varepsilon^{-4}\log\mpt{C_2/\delta}$, then for any fixed $\bm{z}, \bm{z}'\in \mathbb{S}^{d-1}$, with probability at least $1-\delta$, we have
\[\abs{r_0^{(p),m}\mpt{\bm{z},\bm{z}'}-r\mpt{\bm{z},\bm{z}'}}\leq\varepsilon.\]
\end{theorem}
Let $\varepsilon=m^{-1/5}$, we can get the following corollary.
\begin{corollary}\label{prop:initialization of NTK}

There exist some positive absolute constants $C_1>0$ and $C_2\geq 1$, such that if $\delta\in\pt{0,1}$ and $m\geq C_1\pt{\log\mpt{C_2/\delta}}^5$, then for any fixed $\bm{z},\bm{z}'\in \mb{S}^{d-1}$, with probability at least $1-\delta$, we have
\[\abs{r_0^{(p),m}\mpt{\bm{z},\bm{z}'}-r\mpt{\bm{z},\bm{z}'}}=O\mpt{m^{-1/5}}.\]

\end{corollary}

\begin{lemma}[Corollary 5.35 in \cite{vershynin2010introduction}]\label{lem: random matrix base}
Let $\bm{M}$ be an $a\times b$ matrix whose entries are independent standard normal random variables. Then for every $t\geq 0$, with probability at least $1-2\exp\mpt{- t^2/2}$, we have
\[\norm{\bm{M}}_2\leq\sqrt{a}+\sqrt{b}+t.\]
\end{lemma}

According to this lemma, we can directly get
\begin{corollary}[Random matrix]\label{coro: Random matrix}
At initialization, there exists a positive absolute constant $C$, such that if $m\geq C$, then with probability at least $1-\exp\mpt{-\Omega(m)}$, the spectral norm of each matrix satisfies
\[\norm{\bm{A}}_2 = O\mpt{\sqrt{m}},~\norm{\bm{W}^{(l)}_0}_2 = O\mpt{\sqrt{m}},~\norm{\bm{V}^{(l)}_0}_2 = O\mpt{\sqrt{m}}~\text{and}~\norm{\bm{v}}_2 = O\mpt{\sqrt{m}}\quad\text{for}~l\in[L].\]
\end{corollary}

\begin{lemma}\label{lem:up_bound_of_alpha_delta}There exists a positive absolute constant $C$, such that if $m\geq C$, then with probability at least $1-\exp\mpt{-\Omega(m)}$ over the randomness of $\bm{A}$, $\bm{W}^{(l)}_0$, $\bm{V}^{(l)}_0$ and $\bm{v}$, for any $\bm{x}\in\mb{S}^{d-1}$, we have
 \begin{align*}
 \norm{\bm{\alpha}^{(l)}_{0,\bm x}}_{2} =O(1) \quad~\text{and} \quad~\norm{\bm{\delta}^{(l)}_{0,\bm x}}_2=O\mpt{\sqrt m}\qquad~\text{for}~l\in\cl{0,1,\cdots,L}.
 \end{align*}
\end{lemma}

\begin{proof}

We prove by induction that $\norm{\bm{\alpha}^{(l)}_{0,\bm x} }_{2} =O(1)$.

Base case: since $\norm{\bm{x}}_2=1$, with high probability over the random initialization of $\bm{A}$, we have $\norm{\bm{\alpha}^{(0)}_{0,\bm x}}_2= \norm{\bm{A}\bm{x}/\sqrt m}_2\leq \norm{\bm{A}}_2/\sqrt m=O(1)$.
Assume that $\norm{\bm{\alpha}^{(l-1)}_{0,\bm x}}_2 =O(1)$, we can get\begin{align*}
\norm{\bm{\alpha}^{(l)}_{0,\bm x}}_{2} &= \norm{\bm{\alpha}^{(l-1)}_{0,\bm x} +a\frac{1}{\sqrt{m}}\bm{V}^{(l)}_0 \sigma\mpt{\sqrt{\frac{2}{m}}\bm{W}^{(l)}_0 \bm{\alpha}^{(l-1)}_{0,\bm x}}}_{2} \\
&\leq \pt{1 + \frac{\sqrt 2 a}{m}\norm{\bm{V}^{(l)}_0}_{2} \norm{\bm{W}^{(l)}_0}_2 } \norm{\bm{\alpha}^{(l-1)}_{0,\bm x}}_2=O(1)
\end{align*}
with high probability over the random initialization of $\bm{W}^{(l)}_0$ and $\bm{V}^{(l)}_0$.
Therefore, for all $l\in\cl{0,1,\cdots,L}$, we have $\left\|\bm{\alpha}^{(l)}_{0,\bm x} \right\|_{2} =O(1)$.

Next, we prove by induction that $\norm{\bm{\delta}^{(l)}_{0,\bm x}}_{2} =O\mpt{\sqrt m}$. Recall that
\[\bm{\delta}^{(l)}_{0,\bm x}=\pt{\bm{I}_m + \frac{\sqrt 2 a}{m} \bm{V}^{(l+1)} \bm{D}^{(l+1)} \bm{W}^{(l+1)}}^T\bm{\delta}^{(l+1)}_{0,\bm x}.\]

Base case: $\norm{\bm{\delta}^{(L)}_{0,\bm x}}_2 = \norm{ \bm{v}}_2=O\mpt{\sqrt m}$.
Assume that $\norm{\bm{\delta}^{(l+1)}_{0,\bm x}}_2=O\mpt{\sqrt m}$, we can get
\begin{align*}
 \norm{\bm{\delta}^{(l)}_{0,\bm x}}_{2} &=\norm{\pt{\bm{I}_m + \frac{\sqrt 2 a}{m}\bm{W}^{(l+1)}_0 \bm{D}^{(l+1)}_{0,\bm x} \bm{V}^{(l+1)}_0}^T \bm{\delta}^{(l+1)}_{0,\bm x} }_2\\
 &=\pt{1+\frac{\sqrt 2 a}{m} \norm{\bm{W}^{(l+1)}_0}_2 \norm{\bm{V}^{(l+1)}_0}_2 } \norm{\bm{\delta}^{(l+1)}_{0,\bm x}}_2=O\mpt{\sqrt m}
\end{align*}
with probability at least $1-\exp\mpt{-\Omega(m)}$. Therefore, for all $l\in \{0,1,\cdots,L\}$, we have $\norm{\bm{\delta}^{(l)}_{0,\bm x}}_2=O\mpt{\sqrt m}$.

\end{proof}

\begin{lemma}[Johnson-Lindenstrauss Theorem in \cite{dasgupta2003elementary}]\label{lem:Johnson-Lindenstrauss Theorem}
 Let $\bm{h}\in\mb{R}^p$ be a fixed vector, $\bm{W}\in \mathbb{R}^{m\times p}$ be the random matrix with i.i.d. entries $\bm{W}_{i,j}\iid \mathcal{N}(0,1)$, and the vector $\bm{v}\in\mathbb{R}^{m}$ defined as $\bm{v}=\bm{Wh}/\sqrt{m}$. Then for any $\varepsilon\in(0,1)$, with probability at least $1-2\exp\mpt{-\Omega(\varepsilon^2m)}$, we have
 \begin{align*}
 \norm{\bm{h}}_2(1-\varepsilon)\leq \norm{\bm{v}}_2\leq \norm{\bm{h}}_2(1+\varepsilon).
 \end{align*}
\end{lemma}

\begin{lemma}[Lemma 7.1 in \cite{allen2019convergence}]\label{lem: norm_layer_by_layer}
 Let $\bm{h}\in \mathbb{R}^{p}$ be a fixed vector, $\bm{W}\in\mathbb{R}^{m\times p}$ be the random matrix with i.i.d. entries $\bm{W}_{i,j}\iid \mathcal{N}(0,1)$, and the vector $\bm{v}\in\mathbb{R}^{m}$ defined as $\bm{v}=\sigma\mpt{\sqrt{{2}/{m}}\bm{Wh}}$. Then there exists a positive absolute constant $C$, such that for any $\varepsilon\in(0,1)$, with probability of $1-\exp\mpt{-\Omega(\varepsilon^2m)}$, we have
 \begin{align*}
\norm{\bm{h}}_2(1-\varepsilon)\leq \norm{\bm{v}}_2\leq \norm{\bm{h}}_2(1+\varepsilon)
 \end{align*}
when $m\geq C\varepsilon^{-2}$.
\end{lemma}
\paragraph*{Remark:} Lemma \ref{lem: norm_layer_by_layer} is a special case for $L=1$ in Lemma 7.1 of \cite{allen2019convergence}.

\begin{lemma}\label{lem: low_bound_of_alpha_0}There exists a positive absolute constant $C$, such that for any fixed $\bm{z}\in\mb{S}^{d-1}$, with probability at least $1-\exp\mpt{-\Omega(m^{5/6})}$ over the randomness of $\bm{A}$, $\bm{W}^{(l)}_0$ and $\bm{V}^{(l)}_0$, we have
 \begin{align*}
 \norm{\bm{\alpha}^{(l)}_{0,\bm z}}_{2} =\Omega(1)\qquad~\text{for}~l\in\cl{0,1,\cdots,L}
 \end{align*}
when $m$ is greater than the positive constant $C$.
\end{lemma}

\begin{proof}[Proof of Lemma \ref{lem: low_bound_of_alpha_0}]
 We prove by induction that $\norm{\bm{\alpha}^{(l-1)}_{0,\bm z}}_{2} =\Omega(1)$.

 Base case: Since $\norm{\bm{z}}=1$ and $\bm{\alpha}^{(0)}_{0,\bm z}=\sqrt{{1}/{m}}\bm{A}\bm{z}$, by Lemma \ref{lem:Johnson-Lindenstrauss Theorem}, we can get $\norm{\bm{\alpha}^{(0)}_{0,\bm z}}_2=\Omega(1)$ with probability at least $1-\exp\mpt{-\Omega(m)}$.

 Assume that $\norm{\bm{\alpha}^{(l-1)}_{0,\bm z}}_{2} =\Omega(1)$. Denote $\tilde{\bm{\alpha}}^{(l)} = \sigma\mpt{\sqrt{{2}/{m}}\bm{W}^{(l)} \bm{\alpha}^{(l-1)}_{0,\bm z}}$. By Lemma \ref{lem: norm_layer_by_layer}, we have 
 
 \[\norm{\bm{\alpha}^{(l-1)}_{0,\bm z} }_2(1-\varepsilon)\leq \norm{\tilde{\bm{\alpha}}^{(l)}}_2\leq \norm{\bm{\alpha}^{(l-1)}_{0,\bm z} }_2(1+\varepsilon)\]
 
 with probability at least $1-\exp\mpt{-\Omega(\varepsilon^2m)}$ when $m\geq C_1\varepsilon^{-2}$ for some positive constant $C_1$. 

 Denote $\bm{u}^{(l)} = \bm{V}^{(l)}\tilde{\bm{\alpha}}^{(l)}/\sqrt{m}$. By Lemma \ref{lem:Johnson-Lindenstrauss Theorem}, for $\varepsilon\in(0,1)$, we have 
 
 \[\norm{\tilde{\bm{\alpha}}^{(l)}}_2(1-\varepsilon)\leq\norm{\bm{u}^{(l)}}_2\leq\norm{\tilde{\bm{\alpha}}^{(l)}}_2(1+\varepsilon)\]
 
 with probability at least $1-2\exp\mpt{-\Omega(\varepsilon^2m)}$. To sum up,
 \begin{align*}
 \norm{\bm{\alpha}^{(l)}_{0,\bm z} }_2&=\norm{\bm{\alpha}^{(l-1)}_{0,\bm z} +a\bm{u}^{(l)}}_2\geq \abs{\norm{\bm{\alpha}^{(l-1)}_{0,\bm z} }_2-a\norm{\bm{u}^{(l)}}_2}\\
 &\geq \pt{1-a(1+\varepsilon)^2}\norm{\bm{\alpha}^{(l-1)}_{0,\bm z} }_2.
 \end{align*}
 Take $\varepsilon=m^{-1/12}$, since $a< 1$, for sufficient large $m$, with probability at least $1-\exp\mpt{-\Omega(m^{5/6})}$, we can get
 \begin{align*}
 \norm{\bm{\alpha}^{(l)}_{0,\bm z}}_{2}&\geq C_1\norm{\bm{\alpha}^{(l-1)}_{0,\bm z}}_2>C_2
 \end{align*}
 for some positive constants $C_1$ and $C_2$.
 
 Therefore, we have that for all $l\in \{0,1,\cdots,L\}$, $\norm{\bm{\alpha}^{(l)}_{0,\bm z} }_{2}=\Omega(1)$.
 
\end{proof}

\begin{lemma}
 \label{lem:NTK_f0}There exists a positive absolute constant $C$, such that with probability at least $1-\exp\mpt{-\Omega(m)}$, for any $\bm{x}\in\mb{S}^{d-1}$, we have
 \begin{align*}
 \norm{\nabla_{\bm{W}^{(l)}} f^{(p),m}_0(\bm{x}) }_{F} = O(1), \quad \norm{\nabla_{\bm{V}^{(l)}} f^{(p),m}_0(\bm{x}) }_{F} = O(1) \quad l\in\{1,\cdots,L\}
 \end{align*}
when $m$ is greater than the positive constant $C$.
\end{lemma}

\begin{proof}[Proof of Lemma \ref{lem:NTK_f0}]

First of all, because \[\norm{\bm{ab}^T}_F^2=\mr{Tr}\mpt{\bm{ab}^T\bm{ba}^T}=\mr{Tr}\mpt{\bm{a}^T\bm{a}\bm{b}^T\bm{b}}=\norm{\bm{a}}_2^2\norm{\bm{b}}_2^2\]
holds for two vectors $\bm{a}$ and $\bm{b}$, we can easily get
 \begin{align*}
 &\norm{\nabla_{\bm{W}^{(l)}} f^{(p),m}_0(\bm{x}) }_{F} =\frac{\sqrt{2}a}{m} \norm{\pt{\bm{D}_{0,\bm x}^{(l)} \bm{V}^{(l),T}_0 \bm{\delta}_{0,\bm x}^{(l)}}\bm{\alpha}_{0,\bm x}^{(l-1),T} }_{F}\\
 &\qquad=\frac{\sqrt{2}a}{m} \norm{\bm{D}_{0,\bm x}^{(l)} \bm{V}^{(l),T}_0 \bm{\delta}_{0,\bm x}^{(l)}}_2\norm{\bm{\alpha}_{0,\bm x}^{(l-1),T} }_{2}\leq \frac{\sqrt{2} a}{m}\norm{ \bm{V}^{(l)}_0 }_{2} \norm{\bm{\delta}^{(l)}_{0,\bm x}}_{2}\norm{\bm{\alpha}^{(l-1)}_{0,\bm x}}_{2}
 \end{align*}
 and
\begin{align*}
 &\norm{\nabla_{\bm{V}^{(l)}} f^{(p),m}_0(\bm{x}) }_{F} = \frac{\sqrt 2 a}{m}\norm{\bm{\delta}^{(l)}_0 \sigma^T\mpt{\bm{W}^{(l)}_0 \bm{\alpha}^{(l-1)}_0}}_{F}\\
 &\qquad=\frac{\sqrt 2 a}{m}\norm{\bm{\delta}^{(l)}_0}_2\norm{\sigma\mpt{\bm{W}^{(l)}_0 \bm{\alpha}^{(l-1)}_0}}_{2}\leq \frac{\sqrt 2a}{m} \norm{\bm{\delta}^{(l)}_0}_{2} \norm{\bm{W}^{(l)}_0}_{2} \norm{\bm{\alpha}^{(l-1)}_0}_{2}.
\end{align*}
By Corallary \ref{coro: Random matrix}, we know that with probability at least $1-\exp\mpt{-\Omega(m)}$, $\norm{\bm{W}^{(l)}_0}_2 = O\mpt{\sqrt{m}}$ and $\norm{\bm{V}^{(l)}_0}_2 = O\mpt{\sqrt{m}}$ hold when $m$ is greater than some positive constant. Also, by Lemma \ref{lem:up_bound_of_alpha_delta}, we have shown that $\norm{\bm{\alpha}^{(l-1)}_{0,\bm x}}_{2} =O(1)$ and $\norm{\bm{\delta}^{(l)}_{0,\bm x}}_{2} =O\mpt{\sqrt m}$ will hold under the similar conditions. Thus, we have
\begin{align*}
 \norm{\nabla_{\bm{W}^{(l)}} f^{(p),m}_0(\bm{x}) }_{F} = O(1), \quad \norm{\nabla_{\bm{V}^{(l)}} f^{(p),m}_0(\bm{x}) }_{F} = O(1) \quad ~\text{for}~l\in\{1,\cdots,L\}.
\end{align*}
\end{proof}

\subsection{During training}

\begin{lemma}[Corollary 8.4 in \cite{allen2019convergence}]\label{lem: D'_Estimation}
 Suppose $\delta\in[0,O(1)]$ and $\bm{W}_0\in \mathbb{R}^{m\times m}$ is a random matrix with entries drawn i.i.d from $\mathcal{N}(0,1)$. With probability at least $1-\exp\mpt{-\Omega(m\delta^{2/3})}$, the following holds. 
 Fix any vector $\bm{h}\in \mathbb{R}^{m}$ with $\norm{\bm{h}}_2=\Theta(1)$ and for all $\bm{g}'\in \mathbb{R}^{m}$ with $\norm{\bm{g}'}_2\leq \delta$.

 Let $\bm{D}'$ be the diagonal matrix where
 \[(\bm{D}')_{k,k} = \bm{1}\cl{\pt{\sqrt{\tfrac{2}{m}}\bm{W}_0 \bm{h} + \bm{g}'}_k>0} - \bm{1}\cl{\pt{\sqrt{\tfrac{2}{m}}\bm{W}_0 \bm{h}}_k>0}.\] Then, letting $\bm{u}=\bm{D}'\pt{\sqrt{{2}/{m}} \bm{W}_0 \bm{h} +\bm{g}'}$, we have
 \begin{align*}
 \norm{\bm{u}}_0\leq \norm{\bm{D}'}_0 = O(m\delta^{2/3}), \quad \norm{\bm{u}}_2=O(\delta).
 \end{align*}
\end{lemma}

\begin{lemma}\label{lem:Recur_uv}
 Suppose each entry of matrix $\bm{W} \in \mathbb{R}^{a \times b}$ follows $\bm{W}_{ij} \iid\mathcal{N}(0,1)$. Let $c = \max(a,b)$. If $s\geq 0$, then with probability at least $1- \exp\mpt{-s \log c}$, the following holds:
 \begin{equation*}
 \forall \bm{u} \in \mathbb{R}^a, \bm{v} \in \mathbb{R}^b ,\, s.t. \norm{\bm{u}}_0 \norm{\bm{v}}_0 \leq s, \quad\text{we have }\quad \lvert \bm{u}^T \bm{W v} \rvert \leq 9 \sqrt{s \log c } \norm{\bm{u}}_2 \norm{\bm{v}}_2.
 \end{equation*}

\end{lemma}

\begin{proof}
 First of all, it is easy to see that when $s<1 $ or $c=1$, the proposition is trivial. So we only need to consider the result under condition that $s \geq 1$ and $c \geq 2$. 
 
 Note that we aims to prove the inequality holds uniformly for all $\bm{u},\bm{v}$ such that $\norm{\bm{u}}_0, \norm{\bm{v}}_0 \leq s$ at a high probability, we consider the non-zero entris of $\bm{u},\bm{v}$ at first.
 
 Let $A \subseteq [a]$ such that $\abs{A}= \min\cl{a,\fl{s}}$, and let $U_A = \cl{ \bm{u}\in \mathbb{R}^{\abs{A}} : \forall i \notin A, u_i = 0 } $ be a set that contains vectors of which non-zero entries are only located in $A$. In the same way, let $B \subseteq [b] $ such that $\abs{B}=\min\cl{b, \fl{s}}$, and let $V_B = 
 \cl{ \bm{v}\in \mathbb{R}^{\abs{B}} : \forall j \notin B, v_j = 0 } $. Then we have
 \begin{equation*}
 \bm{u}^T \bm{W v} = \sum^{a}_{i=1} \sum^{b}_{j=1} \bm{u}_i \bm{W}_{ij} \bm{v}_j = \sum_{i \in A,j \in B} \bm{u}_i \bm{W}_{ij} \bm{v}_j= \bm{u}^T_A \bm{W}_{AB} \bm{v}_B,
 \end{equation*}
 in which $\bm{u}_A = (\bm{u}_i)_{i \in A}^T, \bm{v}_B = (\bm{v}_j)_{j \in B}^T, \bm{W}_{AB} = (\bm{W}_{ij})_{i \in A, j \in B}$.
 According to the definition of spectral norm, we know that
 \begin{equation*}
 \abs{\bm{u}^T \bm{W v}} = \abs{ \bm{u}^T_A \bm{\bm{W}}_{AB} \bm{\bm{v}}_B } \leq \norm{\bm{u}_A}_2 \norm{\bm{W}_{AB}}_2 \norm{\bm{v}_B} _2.
 \end{equation*}
 
 Now we consider the spectral norm of $\bm{W}_{AB} \in \mathbb{R}^{\abs{A} \times \abs{B}}$. By Lemma \ref{lem: random matrix base}, we know when $t \geq \sqrt{\fl{s}}$, with probability at least $1- 2\exp\mpt{-t^2/2}$, we have $\norm{\bm{W}_{AB}}_2 \leq 3t$. Then we have
 \begin{equation*}
 \forall \bm{u} \in U_A, \forall \bm{v} \in V_B, \quad \lvert \bm{u}^T \bm{W v} \rvert \leq \norm{\bm{u}}_2 \norm{\bm{W}_{AB}}_2 \norm{\bm{v}}_2 \leq 3t \norm{\bm{u}}_2 \norm{\bm{v}}_2.
 \end{equation*}

 Now we consider all possible $A$ and $B$, or to say all possible location of non-zero entries. We know there are $\binom{a}{\abs{A}}$ kinds of $ A$ and $\binom{b}{\abs{B}}$ kinds of $B$ in total. Therefore, with probability at least $1- 2\binom{a}{\abs{A}}\binom{b}{\abs{B}} \exp\mpt{-t^2/2} $, the following proposition holds:
\begin{equation*}
 \forall \bm{u} \in \mathbb{R}^a, \forall \bm{v} \in \mathbb{R}^b, s.t. \norm{\bm{u}}_0,\norm{\bm{v}}_0 \leq s, \quad \text{we have } \lvert \bm{u}^T \bm{W v} \rvert \leq 3t \norm{\bm{u}}_2 \norm{\bm{v}}_2.
 \end{equation*}

With the trivial inequality $\binom{n}{k} \leq n^k $, we have a control for the probability above:
\begin{align*}1-2\binom{a}{\abs{A}}\binom{b}{\abs{B}}\exp\mpt{-t^2/2} \geq1-2a^{\abs{A}}b^{\abs{B}}\exp\mpt{- t^2/2}\geq1-2a^{\fl{s}}b^{\fl{s}}\exp\mpt{- t^2/2}&\\
\qquad\geq 1-2c^{2s}\exp\mpt{- t^2/2}=1-\exp\mpt{-\pt{t^2/2-2s\log c-\log 2}}&.\end{align*}
Finally, let $t=\sqrt{8s\log c} \geq \sqrt{s}$, and then we get the expected result.

\end{proof}

\begin{lemma}\label{lem: 3result_point} 
Let $\tau=O\mpt{{\sqrt{m}}/{(\log m)^3}}$ and $T\subseteq [0,\infty)$. Suppose that $\norm{\bm{W}^{(l)}_t - \bm{W}^{(l)}_0 }_{F}\leq\tau$ and $\norm{\bm{V}^{(l)}_t - \bm{V}^{(l)}_0 }_{F}\leq\tau$ hold for all $t\in T$ and $l\in[L]$. Then there exists a positive absolute constant $C$, such that for any fixed $\bm{z}\in\mb{S}^{d-1}$, with probability at least $1-\exp\mpt{-\Omega(m^{2/3}\tau^{2/3})}$, for all $t\in T$ and $l\in[L]$, we have 
\begin{itemize}
 \item $i)$ $\norm{\bm{g}'_{l,\bm{z}\bm{z}}}_2 = O\mpt{{\tau}/{\sqrt{m}}}$;
 \item $ii)$ $\norm{\bm{D}^{(l)'}_{\bm z \bm z}}_0 =O\mpt{m^{2/3}\tau^{2/3}}$ and $\norm{\bm{D}^{(l)'}_{\bm z \bm z} \bm{W}^{(l)}_t \bm{\alpha}^{(l-1)}_{t,\bm z} }_2=O\mpt{{\tau}}$;
 \item $iii)$ $\norm{\Delta\bm{\alpha}^{(l)}_{\bm z \bm z}}_2=O\mpt{{\tau}/{\sqrt{m}}}$.
\end{itemize}
when $m$ is greater than the positive constant $C$. 
\end{lemma}

\begin{proof}[Proof of Lemma \ref{lem: 3result_point}]
We have shown that $\norm{\bm{W}^{(l)}_0}_2 = O(\sqrt{m})$ and $\norm{\bm{V}^{(l)}_0}_2 = O(\sqrt{m})$ hold with probability at least $1-\exp\mpt{-\Omega(m)}$. Combine with $\norm{\Delta\bm{W}^{(l)}}_F\leq\tau$ and $\norm{\Delta\bm{V}^{(l)}}_F\leq\tau$, we can get 
\[\norm{\bm{W}^{(l)}_t}_2 = O(\sqrt{m})\qquad~\text{and}\qquad~\norm{\bm{V}^{(l)}_t}_2 = O(\sqrt{m}).\]

Since $\bm{A}$ does not change during the training, we can easily check that $\norm{\Delta\bm{\alpha}^{(0)}_{\bm{z},\bm{z}}}_2=\norm{\bm{0}}_2=0=O\mpt{{\tau}/{\sqrt{m}}}$, which means that $iii)$ holds for $l=0$. Then it only needs to be proven that
\[iii)~\text{holds for}~l=k\quad\Longrightarrow\quad~\text{Lemma holds for}~l=k+1.\]
Now we assume that $iii)$ holds for $l=k\in\{0,1,\cdots,L-1\}$, then with probability at least $1-\exp\mpt{-\Omega(m)}$, we have
\[\norm{\bm{\alpha}^{(k)}_{t,\bm z}}_2=\norm{\bm{\alpha}^{(k)}_{0,\bm z}+\Delta\bm{\alpha}^{(k)}_{\bm z \bm z}}_2\leq\norm{\bm{\alpha}^{(k)}_0}_2+\norm{\Delta\bm{\alpha}^{(k)}_{\bm z \bm z}}_2=O(1)+O\mpt{\tau/\sqrt{m}}=O(1).\]
For $i)$, we can get
\begin{align*}
 \bm{g}_{k+1,\bm z \bm z}'&= \sqrt{\frac{2}{m}}\pt{\bm{W}^{(k+1)}_t \bm{\alpha}^{(k)}_{t,\bm z} -\bm{W}^{(k+1)}_0 \bm{\alpha}^{(k)}_{0,\bm z}}=\sqrt{\frac{2}{m}}\pt{ \Delta\bm{W}^{(k+1)} \bm{\alpha}^{(k)}_{t,\bm z} + \bm{W}^{(k+1)}_0 \Delta\bm{\alpha}^{(k)}_{\bm z \bm z}}, 
 \end{align*}
which can lead to
 \begin{align*}
 \norm{\bm{g}_{k+1,\bm z \bm z}'}_2&\leq\sqrt{\frac{2}{m}}\pt{ \norm{\Delta\bm{W}^{(k+1)}}_2\norm{\bm{\alpha}^{(k)}_{t,\bm z}}_2 + \norm{\bm{W}^{(k+1)}_0}_2 \norm{\Delta\bm{\alpha}^{(k)}_{\bm z \bm z}}_2}\\
 &\leq\sqrt{\frac{2}{m}}\pt{\tau\cdot O(1)+O\mpt{\sqrt{m}}O\mpt{\tau/\sqrt{m}}}\leq O\mpt{\frac{\tau}{\sqrt m}}.
 \end{align*}
Then by Lemma \ref{lem: D'_Estimation} and taking $\bm{W}_0=\bm{W}^{(k+1)}_0$, $\bm{h}=\bm{\alpha}^{(k)}_{0,\bm z}$, $\bm{g}'=\bm{g}'_{k+1}(\bm z)$ and $\delta=\Theta\mpt{{\tau}/{\sqrt{m}}}\leq O\mpt{(\log m)^{-3}}$, we can get $ii)$ holds for $l=k+1$ with probability at least $1-\exp\mpt{-\Omega(m^{2/3}\tau^{2/3})}$ since we have shown that $\norm{\bm{h}}_2=\norm{\bm{\alpha}^{(k)}_{0,\bm z}}_2=\Theta(1)$ in Lemmas \ref{lem:up_bound_of_alpha_delta} and \ref{lem: low_bound_of_alpha_0}.

As for $iii)$, it is easy to check that
\begin{align*}
\Delta\bm{\alpha}^{(k+1)}_{\bm z \bm z}
&=\Delta\bm{\alpha}^{(k)}_{\bm z \bm z}+\frac{\sqrt{2}a}{m}\Big[\bm{V}^{(k+1)}_t\bm{D}^{(k+1)'}_{\bm z \bm z}\bm{W}^{(k+1)}_t\bm{\alpha}^{(k)}_{t,\bm z}+\Delta\bm{V}^{(k+1)}\bm{D}^{(k+1)}_{0,\bm z}\bm{W}^{(k+1)}_t\bm{\alpha}^{(k)}_{t,\bm z}
\\&\qquad\qquad\qquad\qquad\qquad\qquad+\bm{V}^{(k+1)}_0\bm{D}^{(k+1)}_{0,\bm z}\pt{\bm{W}^{(k+1)}_t\bm{\alpha}^{(k)}_{t,\bm z}-\bm{W}^{(k+1)}_0\bm{\alpha}^{(k)}_{0,\bm z}}\Big].
\end{align*}
We have shown that, with probability at least $1-\exp\mpt{-\Omega(m^{2/3}\tau^{2/3})}$, $i)$ $ii)$ hold for $l=k+1$, i.e.
\begin{align*}
\begin{gathered}
 \norm{\bm{D}^{(k+1)'}_{\bm z \bm z} \bm{W}^{(k+1)}_t \bm{\alpha}^{(k)}_{t,\bm z} }_2=O\mpt{{\tau}};\\
 \norm{\bm{W}^{(k+1)}_t\bm{\alpha}^{(k)}_{t,\bm z}-\bm{W}^{(k+1)}_0\bm{\alpha}^{(k)}_{0,\bm z}}_2=\sqrt{\frac{m}{2}}\norm{\bm{g}'_{k+1}}_2= O(\tau),
\end{gathered}
\end{align*}
which can lead to $\norm{\Delta\bm{\alpha}^{(k+1)}_{\bm z \bm z}}_2= O\mpt{\tau/\sqrt{m}}$. 

Thus, we finish the proof.

\end{proof}

\begin{lemma}\label{lem: Delta_delta_point}
Let $\tau=O\mpt{{\sqrt{m}}/{(\log m)^3}}$ and $T\subseteq [0,\infty)$. Suppose that $\norm{\bm{W}^{(l)}_t - \bm{W}^{(l)}_0 }_{F}\leq\tau$ and $\norm{\bm{V}^{(l)}_t - \bm{V}^{(l)}_0 }_{F}\leq\tau$ hold for all $t\in T$ and $l\in[L]$. Then there exists a positive absolute constant $C$, such that for any fixed $\bm{z}\in\mb{S}^{d-1}$, with probability at least $1-\exp\mpt{-\Omega(m^{2/3}\tau^{2/3})}$, for all $t\in T$ and $l\in[L]$, we have 
\begin{align*}
\norm{\Delta\bm{\delta}^{(l)}_{\bm z \bm z}}_2= O\mpt{m^{1/3}\tau^{1/3}\sqrt{\log m}},
\end{align*}
when $m$ is greater than the positive constant $C$. 
\end{lemma}

\begin{proof}[Proof of Lemma \ref{lem: Delta_delta_point}]
We inductively prove this lemma. 

Base case: $\norm{\Delta\bm{\delta}^{(L)}_{\bm z \bm z}}_2 = 0$ since $\bm{v}$ is fixed during the training process.

Assume that this lemma holds for $l+1$, then with probability at least $1-\exp\mpt{-\Omega(m)}$, we have
\[\norm{\bm{\delta}^{(l+1)}_{t,\bm z}}_2=\norm{\Delta\bm{\delta}^{(l+1)}_{\bm z \bm z}+\bm{\delta}^{(l+1)}_{0,\bm z}}_2\leq\norm{\Delta\bm{\delta}^{(l+1)}_{\bm z \bm z}}_2+\norm{\bm{\delta}^{(l+1)}_{0,\bm z}}_2\leq O(\sqrt{ m})\]
because of Lemma \ref{lem:up_bound_of_alpha_delta}. Moreover, it is easy to check that
\begin{align*}
\Delta\bm{\delta}^{(l)}_{\bm z \bm z}
&=\Delta\bm{\delta}^{(l+1)}_{\bm z \bm z}+\tfrac{\sqrt{2}a}{m}\Big[\bm{W}^{(l+1),T}_0\bm{D}^{(l+1)'}_{\bm z \bm z}\bm{V}^{(l+1),T}_0\bm{\delta}^{(l+1)}_{0,\bm z}+\Delta\bm{W}^{(l+1),T}\bm{D}^{(l+1)}_{t,\bm z}\bm{V}^{(l+1),T}_0\bm{\delta}^{(l+1)}_{0,\bm z}
\\&\qquad\qquad\qquad~+\bm{W}^{(l+1),T}_t\bm{D}^{(l+1)}_{t,\bm z}\Delta\bm{V}^{(l+1),T} \bm{\delta}^{(l+1)}_{0,\bm z} + \bm{W}^{(l+1),T}_t\bm{D}^{(l+1)}_{t,\bm z}\bm{V}^{(l+1),T}_t \Delta\bm{\delta}^{(l+1)}_{\bm z \bm z}\Big].
\end{align*}
Let us denote the four terms within the square brackets, excluding the factor `$\sqrt 2a/m$' outside the brackets, as $\bm{u}_1$ to $\bm{u}_4$ respectively. First of all, it is easy to check that, with probability at least $1-\exp\mpt{-\Omega(m)}$, we have
\begin{align*}
\begin{gathered}
\norm{\bm{u}_2}_2\leq \tau \cdot1\cdot O\mpt{\sqrt{m}}\cdot O\mpt{\sqrt{m}}=O\mpt{\tau \cdot m}\leq O\mpt{m\cdot m^{1/3}\tau^{1/3}\sqrt{\log m}};\\
\norm{\bm{u}_3}_2\leq O\mpt{\sqrt{m}} \cdot1\cdot \tau \cdot O\mpt{\sqrt{m}}=O\mpt{\tau \cdot m}\leq O\mpt{m\cdot m^{1/3}\tau^{1/3}\sqrt{\log m}};\\
\norm{\bm{u}_4}_2\leq O\mpt{\sqrt{m}}\cdot 1\cdot O\mpt{\sqrt{m}}\cdot O\mpt{m^{1/3}\tau^{1/3}\sqrt{\log m}}=O\mpt{m\cdot m^{1/3}\tau^{1/3}\sqrt{\log m}}.
\end{gathered}
\end{align*}
As for $\bm{u}_1$, if ${\bm{\delta}^{(l+1)}_{0,\bm z}}=\bm{0}$ or $\norm{\bm{D}^{(l+1)'}_{\bm z \bm z}}_0=0$, we have $\norm{\bm{u}_1}_2=0$. Therefore, we consider the case where ${\bm{\delta}^{(l+1)}_{0,\bm z}}\not=\bm{0}$ and $\norm{\bm{D}^{(l+1)'}_{\bm{z}\bm{z}}}_{0}\geq1$. Denote $\tilde{\bm{\delta}}={\bm{\delta}^{(l+1)}_{0,\bm z}}/\norm{\bm{\delta}^{(l+1)}_{0,\bm z}}_2$ for $\bm{\delta}^{(l+1)}_{0,\bm z}\in\mb{R}^{m}\backslash\{\bm{0}\}$, we can get
\begin{align*}
\norm{\bm{u}_1}_2&\leq \norm{\bm{W}^{(l+1)}_0}_2\norm{\bm{D}^{(l+1)'}_{\bm z \bm z}\bm{V}^{(l+1)}_0\tilde{\bm{\delta}}}_2\norm{\bm{\delta}^{(l+1)}_{0,\bm z}}_2\leq O\mpt{m}\norm{\bm{D}^{(l+1)'}_{\bm z \bm z}\bm{V}^{(l+1)}_0\tilde{\bm{\delta}}}_2.
\end{align*}
Using the randomness of $\bm{V}^{(l+1)}_0$, for any fixed $\tilde{\bm{\delta}}$, we have $\bm{V}^{(l+1)}_0\tilde{\bm{\delta}}\sim\mathcal{N}(\bm{0},\bm{I}_m)$. Thus, by Lemma \ref{lem:Recur_uv} and taking $s=\Theta(m^{2/3}\tau^{2/3})$, with probability at least $1-\exp\mpt{-\Omega(m^{2/3}\tau^{2/3})}$, we can get
\begin{align*}\norm{\bm{D}^{(l+1)'}_{\bm z \bm z}\bm{V}^{(l+1)}_0\tilde{\bm{\delta}}}_2&=\sup_{\bm{u}\in\mathbb{S}^{m-1}}\norm{\bm{u}^T\bm{D}^{(l+1)'}_{\bm z \bm z}\bm{V}^{(l+1)}_0\tilde{\bm{\delta}}}_2=\sup_{\bm{u}\in\mathbb{S}^{m-1}}\norm{\pt{\bm{D}^{(l+1)'}_{\bm z \bm z}\bm{u}}^T\bm{V}^{(l+1)}_0\tilde{\bm{\delta}}\cdot 1}_2\\
&\leq O\mpt{\sqrt{m^{2/3}\tau^{2/3}\log m}}=O\mpt{{m^{1/3}\tau^{1/3}\sqrt{\log m}}},
\end{align*}
because of $\norm{\bm{D}^{(l+1)'}_{\bm z \bm z}\bm{u}}_0\leq\norm{\bm{D}^{(l+1)'}_{\bm z \bm z}}_0\leq \Theta(m^{2/3}\tau^{2/3})$ and $\norm{1}_0=1\leq \norm{\bm{D}^{(l+1)'}_{\bm z \bm z}}_0$.

Combining the above discussions, we can conclude that $\norm{\Delta\bm{\delta}^{(l)}_{\bm z \bm z}}_2\leq O\mpt{m^{1/3}\tau^{1/3}\sqrt{\log m}}$.

\end{proof}

\begin{lemma}\label{lem: gamma_eta_t_vs_0_point}Fix $l\in [L]$ and let $\tau=O\mpt{{\sqrt{m}}/{(\log m)^3}}$, $T\subseteq [0,\infty)$. Suppose that $\norm{\bm{W}^{(l)}_t - \bm{W}^{(l)}_0 }_{F}\leq\tau$ and $\norm{\bm{V}^{(l)}_t - \bm{V}^{(l)}_0 }_{F}\leq\tau$ hold for all $t\in T$, then there exists a positive absolute constant $C$, such that for any fixed $\bm{z}\in\mb{S}^{d-1}$, with probability at least $1-\exp\mpt{-\Omega(m^{2/3}\tau^{2/3})}$ over the randomness of $\bm{W}^{(l)}_0$ and $\bm{V}^{(l)}_0$, for all $t\in T$, we have
 \begin{align*}
 \begin{gathered}
 \norm{\Delta\bm{\gamma}^{(l)}_{\bm z \bm z}}_{2}= O\mpt{m^{-1/6}\tau^{1/3}\sqrt{\log m}}, \quad~ \norm{\Delta\bm{\eta}^{(l)}_{\bm z \bm z}}_{2} = O\mpt{\frac{\tau}{{m}}};\\
 \norm{\bm{\gamma}^{(l)}_{t,\bm z}}_2=O(1),\qquad~ \norm{\bm{\eta}^{(l)}_{t,\bm z}}_2 =O(1/\sqrt{m})
 \end{gathered}
 \end{align*}
when $m$ is greater than the positive constant $C$. 
\end{lemma}

\begin{proof}[Proof of Lemma \ref{lem: gamma_eta_t_vs_0_point}]
First of all, we have
\begin{align*}
&\Delta\bm{\gamma}^{(l)}_{\bm z \bm z}=\frac{\sqrt{2}}{m}\pt{\bm{D}^{(l)}_{t,\bm z}\bm{V}^{(l),T}_t\bm{\delta}^{(l)}_{t,\bm z}-\bm{D}^{(l)}_{0,\bm z}\bm{V}^{(l),T}_0\bm{\delta}^{(l)}_{0,\bm z}}\\
&\qquad=\frac{\sqrt{2}}{m}\pt{\bm{D}^{(l)'}_{\bm z \bm z}\bm{V}^{(l),T}_0\bm{\delta}^{(l)}_{0,\bm z}+\bm{D}^{(l)}_{t,\bm z}\Delta\bm{V}^{(l),T}\bm{\delta}^{(l)}_{0,\bm z}+\bm{D}^{(l)}_{t,\bm z}\bm{V}^{(l),T}_t\Delta\bm{\delta}^{(l)}_{\bm z \bm z}}.
\end{align*}
Using the similar proof technique as the previous lemma, we can establish that with probability at least $1-\exp\mpt{-\Omega(m^{2/3}\tau^{2/3})}$, we have
\[\norm{\bm{D}^{(l)'}_{\bm z \bm z}\bm{V}^{(l),T}_0\bm{\delta}^{(l)}_{0,\bm z}}_2=O\mpt{{m^{1/3}\tau^{1/3}\sqrt{\log m}}}.\]
According to Corollary \ref{coro: Random matrix}, Lemma \ref{lem:up_bound_of_alpha_delta} and Lemma \ref{lem: Delta_delta_point} we can get
\begin{align*}
\norm{\bm{D}^{(l)}_{t,\bm z}\Delta\bm{V}^{(l),T}\bm{\delta}^{(l)}_{0,\bm z}}_2=O\mpt{\tau\sqrt{m}};\qquad~\norm{\bm{D}^{(l)}_{t,\bm z}\bm{V}^{(l),T}_t\Delta\bm{\delta}^{(l)}_{\bm z \bm z}}_2=O\mpt{m^{5/6}\tau^{1/3}\sqrt{\log m}}.
\end{align*}
Thus, we can get $\norm{\Delta\bm{\gamma}^{(l)}_{\bm z \bm z}}_{2}= O\mpt{m^{-1/6}\tau^{1/3}\sqrt{\log m}}$.

As for $\norm{\Delta\bm{\eta}^{(l)}_{\bm z \bm z}}_{2}$, we can similarly get
\begin{align*}
\norm{\Delta\bm{\eta}^{(l)}_{\bm z \bm z}}_2&=\frac{\sqrt{2}}{m}\norm{\bm{D}^{(l)'}_{\bm z \bm z}\bm{W}^{(l)}_t\bm{\alpha}^{(l-1)}_{t,\bm z}+\bm{D}^{(l)}_{0,\bm z}\Delta\bm{W}^{(l)}\bm{\alpha}^{(l-1)}_{t,\bm z}+\bm{D}^{(l)}_{0,\bm z}\bm{W}^{(l)}_0\Delta\bm{\alpha}^{(l-1)}_{\bm z \bm z} }_2\\
&\leq \frac{\sqrt{2}}{m}\big(O\mpt{\tau}+O\mpt{\tau}+O\mpt{\tau}\big)=O\mpt{\frac{\tau}{m}}
\end{align*}
according to Corollary \ref{coro: Random matrix}, Lemma \ref{lem:up_bound_of_alpha_delta} and Lemma \ref{lem: 3result_point}.

With the above results, we can easily get 
\begin{align*}
\begin{gathered}
 \norm{\bm{\gamma}^{(l)}_{0,\bm z}}_2=\frac{\sqrt{2}}{m}\norm{\bm{D}^{(l)}_{0,\bm z}\bm{V}^{(l),T}_0\bm{\delta}^{(l)}_{0,\bm z}}_2=O\mpt{1},\quad~ \norm{\bm{\eta}^{(l)}_{0,\bm z}}_2=\frac{\sqrt{2}}{m}\norm{\bm{D}^{(l)}_{0,\bm z}\bm{W}^{(l)}_0\bm{\alpha}^{(l-1)}_{0,\bm z}}_2=O\mpt{\tfrac{1}{\sqrt{m}}};\\
\norm{\bm{\gamma}^{(l)}_{t,\bm z}}_2\leq\norm{\bm{\gamma}^{(l)}_{0,\bm z}}_2+\norm{\Delta\bm{\gamma}^{(l)}_{\bm z \bm z}}_2=O(1),\quad\norm{\bm{\eta}^{(l)}_{t,\bm z}}_2\leq\norm{\bm{\eta}^{(l)}_0}_2+\norm{\Delta\bm{\eta}^{(l)}_{\bm z \bm z}}_2=O\mpt{\tfrac{1}{\sqrt{m}}}
 \end{gathered}
\end{align*}
since $\tau=O\mpt{{\sqrt{m}}/{(\log m)^3}}$.

\end{proof}

\begin{lemma}
 \label{lem:NTK_ft}
Let $\tau=O\mpt{{\sqrt{m}}/{(\log m)^3}}$ and $T\subseteq [0,\infty)$. Suppose that $\norm{\bm{W}^{(l)}_t - \bm{W}^{(l)}_0 }_{F}\leq\tau$ and $\norm{\bm{V}^{(l)}_t - \bm{V}^{(l)}_0 }_{F}\leq\tau$ hold for all $t\in T$ and $l\in[L]$. Then there exists a positive absolute constant $C$, such that for any fixed $\bm{z}\in\mb{S}^{d-1}$, with probability at least $1-\exp\mpt{-\Omega(m^{2/3}\tau^{2/3})}$, for all $l\in[L]$, we have 
\begin{align*}
\begin{gathered}
 \sup_{t\in T}\norm{\nabla_{\bm{W}^{(l)}} f^{(p),m}_t(\bm{z}) - \nabla_{\bm{W}^{(l)}} f^{(p),m}_0(\bm{z})}_{F}= O\mpt{m^{-1/6}\tau^{1/3} \sqrt{\log m} };\\
 \sup_{t\in T}\norm{\nabla_{\bm{V}^{(l)}} f^{(p),m}_t(\bm{z}) - \nabla_{\bm{V}^{(l)}} f^{(p),m}_0(\bm{z})}_{F}= O\mpt{m^{-1/6}\tau^{1/3}\sqrt{\log m}},
 \end{gathered}
 \end{align*}
 when $m$ is greater than the positive constant $C$. 
\end{lemma}

\begin{proof}[Proof of Lemma \ref{lem:NTK_ft}]According to Eq.(\ref{eq: nabla_W_V_f}), we have
\begin{align*}
 &\norm{\nabla_{\bm{W}^{(l)}} f^{(p),m}_t(\bm{z})- \nabla_{\bm{W}^{(l)}} f^{(p),m}_0(\bm{z})}_{F} = \norm{ a \bm{\gamma}^{(l)}_{t,\bm z} \bm{\alpha}^{(l-1),T}_{t,\bm z} - a \bm{\gamma}^{(l)}_{0,\bm z}\bm{\alpha}^{(l-1),T}_{0,\bm z}}_{F}\\
 &\qquad=a\norm{\bm{\gamma}^{(l)}_{t,\bm z}\Delta\bm{\alpha}^{(l-1),T}_{\bm z \bm z}+\Delta\bm{\gamma}^{(l)}_{\bm z \bm z}\bm{\alpha}^{(l-1),T}_{0,\bm z}}_F\leq a\norm{\bm{\gamma}^{(l)}_{t,\bm z}\Delta\bm{\alpha}^{(l-1),T}_{\bm z \bm z}}_F+a\norm{\Delta\bm{\gamma}^{(l)}_{\bm z \bm z}\bm{\alpha}^{(l-1),T}_{0,\bm z}}_F\\
 &\qquad= a\norm{\bm{\gamma}^{(l)}_{t,\bm z}}_2 \norm{\Delta\bm{\alpha}^{(l-1)}_{\bm z \bm z}}_{2} + a\norm{\Delta\bm{\gamma}^{(l)}_{\bm z \bm z}}_2\norm{\bm{\alpha}^{(l-1)}_{0,\bm z}}_2 \leq O\mpt{m^{-1/6}\tau^{1/3} \sqrt{\log m}}
 \end{align*}
according to Lemma \ref{lem: gamma_eta_t_vs_0_point}, \ref{lem: 3result_point} $iii)$ and \ref{lem:up_bound_of_alpha_delta}. Similarly, we can also get
 \begin{align*}
 &\norm{\nabla_{\bm{V}^{(l)}} f^{(p),m}_t(\bm{z}) - \nabla_{\bm{V}^{(l)}} f^{(p),m}_0(\bm{z})}_{F} = \norm{a \bm{\delta}^{(l)}_{t,\bm z} \bm{\eta}^{(l),T}_{t,\bm z} - a \bm{\delta}^{(l)}_{0,\bm z} \bm{\eta}^{(l),T}_{0,\bm z}}_{F}\\
 &\qquad\leq a\norm{\bm{\eta}^{(l)}_{t,\bm z}}_2 \norm{\Delta\bm{\delta}^{(l)}_{\bm z \bm z}}_{2} + a\norm{\Delta\bm{\eta}^{(l)}_{\bm z \bm z}}_2 \norm{\bm{\delta}^{(l)}_{0,\bm z}}_2\leq O\mpt{m^{-1/6}\tau^{1/3} \sqrt{\log m}}
 \end{align*}
according to Lemma \ref{lem: gamma_eta_t_vs_0_point}, \ref{lem: Delta_delta_point} and \ref{lem:up_bound_of_alpha_delta}.

Thus, we finish the proof.

\end{proof}

\begin{proposition}
 \label{prop:During training}
 Fix $ \bm{z},\bm{z}' \in \mathbb{S}^{d-1}$ and let $\delta\in(0,1)$, $T\subseteq[0,\infty)$. Suppose that $\norm{\bm{W}^{(l)}_t - \bm{W}^{(l)}_0}_{F}= O(m^{1/4})$ and $\norm{\bm{V}^{(l)}_t - \bm{V}^{(l)}_0 }_{F}= O(m^{1/4})$ hold for all $t\in T$ and $l\in[L]$. Then there exist some positive absolute constants $C_1>0$ and $C_2\geq 1$, such that with probability at least $1-\delta$, we have
 \[\sup_{t\in T}\abs{r_t^{(p),m}(\bm{z},\bm{z}') - r_0^{(p),m}(\bm{z},\bm{z}')}= O \mpt{m^{-\frac{1}{12}}\sqrt{\log m}},~\text{when}~m\geq C_1\pt{\log(C_2/\delta)}^{6/5}.\]
\end{proposition}

\begin{proof}[Proof of Proposition \ref{prop:During training}]
By Lemma \ref{lem:NTK_ft} (choose parameter $\tau = \Theta(m^{1/4})$), Lemma \ref{lem:NTK_f0} and
\begin{align*}
 \norm{ \nabla_{\bm{W}^{(l)}} f^{(p),m}_t(\bm{z}')}_{F}&\leq \norm{ \nabla_{\bm{W}^{(l)}} f^{(p),m}_0(\bm{z}')}_{F}+\norm{ \nabla_{\bm{W}^{(l)}} f^{(p),m}_t(\bm{z}')- \nabla_{\bm{W}^{(l)}} f^{(p),m}_0(\bm{z}') }_{F};\\
 \norm{ \nabla_{\bm{V}^{(l)}} f^{(p),m}_t(\bm{z}')}_{F} &\leq \norm{ \nabla_{\bm{V}^{(l)}} f^{(p),m}_0(\bm{z}')}_{F}+\norm{ \nabla_{\bm{V}^{(l)}} f^{(p),m}_t(\bm{z}')- \nabla_{\bm{V}^{(l)}} f^{(p),m}_0(\bm{z}') }_{F},
\end{align*}
with probability at least $1-\exp(-\Omega(m^{5/6}))$, we have
\begin{align*}
&\abs{\ag{\nabla_{\bm{W}^{(l)}} f^{(p),m}_{t}(\bm{z}) , \nabla_{\bm{W}^{(l)}} f^{(p),m}_{t}(\bm{z}')}-\ag{\nabla_{\bm{W}^{(l)}} f^{(p),m}_{0}(\bm{z}) , \nabla_{\bm{W}^{(l)}} f^{(p),m}_{0}(\bm{z}')}}\\
&\qquad\qquad\qquad~~\leq \norm{ \nabla_{\bm{W}^{(l)}} f^{(p),m}_0(\bm{z})}_{F} \norm{ \nabla_{\bm{W}^{(l)}} f^{(p),m}_t(\bm{z}') - \nabla_{\bm{W}^{(l)}} f^{(p),m}_0(\bm{z}')}_{F}\\
&\qquad\qquad\qquad\qquad\qquad\qquad~~+\norm{ \nabla_{\bm{W}^{(l)}} f^{(p),m}_t(\bm{z}')}_{F} \norm{ \nabla_{\bm{W}^{(l)}} f^{(p),m}_t(\bm{z})- \nabla_{\bm{W}^{(l)}} f^{(p),m}_0(\bm{z}) }_{F}\\
&\qquad\qquad\qquad~~ \leq O(1)\cdot O\mpt{m^{-\frac{1}{12}}\sqrt{\log m}}+O(1)\cdot O\mpt{m^{-\frac{1}{12}}\sqrt{\log m}}\leq O\mpt{m^{-\frac{1}{12}}\sqrt{\log m}}
\end{align*}
and similarly have
\begin{align*}\abs{\ag{\nabla_{\bm{V}^{(l)}} f^{(p),m}_{t}(\bm{z}) , \nabla_{\bm{V}^{(l)}} f^{(p),m}_{t}(\bm{z}')}-\ag{\nabla_{\bm{V}^{(l)}} f^{(p),m}_{0}(\bm{z}) , \nabla_{\bm{V}^{(l)}} f^{(p),m}_{0}(\bm{z}')}}\\
\leq O\mpt{m^{-\frac{1}{12}}\sqrt{\log m}}\end{align*}
for all $l\in[L]$ and $t\in T$ when $m$ is greater than some positive absolute constant $C$. Combine with Eq.(\ref{eq: RNK_formulate}), with probability at least $1-\exp(-\Omega(m^{5/6}))$, we can get
\begin{align*}\sup_{t\in T}\abs{r_t^{(p),m}(\bm{z},\bm{z}') - r_0^{(p),m}(\bm{z},\bm{z}')}= O \mpt{m^{-\frac{1}{12}}\sqrt{\log m}}.
\end{align*}
Also, it is easy to check that there exist some positive absolute constants $C_1>0$ and $C_2\geq 1$ such that $C_1\pt{\log\mpt{C_2/\delta}}^{6/5}\geq C$ holds for $\delta\in(0,1)$ and when $m\geq C_1\pt{\log\mpt{C_2/\delta}}^{6/5}$, we have $1-\exp\mpt{-\Omega\mpt{m^{5/6}}} \geq 1-\delta$.


\end{proof}

\subsection{Lazy Regime}

\begin{lemma}
\label{lem: smallest eigenvalue leads to fast convergence}Let $\delta\in(0,1)$ and $t\geq 0$. Suppose that $\norm{\bm{W}^{(p,l)}_s - \bm{W}^{(p,l)}_0}_F = O(m^{1/4})$ and $\norm{\bm{V}^{(p,l)}_s - \bm{V}^{(p,l)}_0}_F = O(m^{1/4})$ hold for all $ s \in [0,t] $, $l\in [L]$ and $p\in[2]$. 
Then there exists a polynomial $\poly(\cdot)$, such that when $m\geq\poly\mpt{n,\lambda_0^{-1},\log(1/\delta)}$, 
with probability at least $1-\delta$, for all $s\in [0,t]$, we have
\[\norm{\bm{u}(s)}_2^{2}\leq \exp\mpt{- \frac{\lambda_{0}}{n}s}\norm{\bm{u}(0)}_2^{2} = \exp\mpt{-\frac{\lambda_{0}}{n}s} \|\bm{y}\|_2^{2},\]
\end{lemma} 
where $\bm{u}(t):=f_t^m(\bm{X})-\bm{y}$.
\begin{proof}Denote $\tilde{\lambda}_0(s)=\lambda_{\min}\big(r_s^m(\bm{X},\bm{X})\big)$. By Weyl's inequality, we can get
 \begin{align*}
&\abs{\tilde{\lambda}_0(s)-\lambda_0}\leq \norm{ r_s^m(\bm{X},\bm{X}) - r(\bm{X},\bm{X}) }_2\leq \norm{ r_s^m(\bm{X},\bm{X}) - r(\bm{X},\bm{X}) }_F
 \\&\qquad\leq\norm{ r_s^m(\bm{X},\bm{X}) - r_0^m(\bm{X},\bm{X}) }_F+\norm{ r_0^m(\bm{X},\bm{X}) - r(\bm{X},\bm{X}) }_F
 \\&\qquad\leq\frac12\sum_{p=1}^2\bk{\sum^n_{i,j=1} \abs{r_s^{(p),m}(\bm{x}_i,\bm{x}_j) - r_0^{(p)m}(\bm{x}_i,\bm{x}_j)} + \sum^n_{i,j=1} \abs{r_0^{(p),m}(\bm{x}_i,\bm{x}_j) - r(\bm{x}_i,\bm{x}_j)}}.
 \end{align*}
According to Proposition \ref{prop:During training} and Corollary \ref{prop:initialization of NTK}, for $\delta_0={\delta}/(2n^2)$, with probability at least $1-2n^2\delta_0=1-\delta$, we can get
\[\abs{\tilde{\lambda}_0(s)-\lambda_0}\leq n^2 \cdot O\mpt{m^{-\frac{1}{12}} \sqrt{\log m}} + n^2\cdot O\mpt{m^{-0.2}} \leq n^2\cdot O\mpt{m^{-\frac1{15}}}\leq \frac{\lambda_0}{2}~\text{for all}~s\in[0,t]\]
when $m\geq C_1\bk{\pt{n^{2}\lambda_0^{-1}}^{15}+\pt{\log\mpt{C_2n^2/\delta}}^5}$ for some positive absolute constants $C_1>0$ and $C_2\geq 1$. This implies that $\tilde{\lambda}_0(s)\geq \lambda_0/2$ holds for all $s\in[0,t]$. Then we have
 \begin{equation*}
 \frac{\mr{d}}{\mr{d} s}\norm{\bm{u}(s)}^{2}_2=- \frac{2}{n}\bm{u}(s)^T K_{s}(\bm{X},\bm{X}) \bm{u}(s)\leq -\frac{\lambda_{0}}{n} \norm{\bm{u}(s)}_{2}^{2}
 \end{equation*}
 and thus
 \begin{equation*}
 \frac{\mr{d}}{\mr{d} s}\mpt{\exp\mpt{\tfrac{\lambda_{0}}{n}s}\norm{ \bm{u}(s)}^{2}_2}=\exp\mpt{\frac{\lambda_{0}}{n}s}\pt{\frac{\lambda_{0}}{n}\lVert \bm{u}(s)\rVert^{2}_2+\frac{\mr{d}\lVert \bm{u}(s)\rVert_2^{2}}{\mr{d} s}}\leq 0.
 \end{equation*}
 Thus, with probability at least $1-\delta$, we can get $\exp\mpt{{\lambda_{0}s}/{n}}\lVert \bm{u}(s)\rVert^{2}_2\leq\|\bm{u}(0)\|_2^2 =\lVert \bm{y} \rVert^{2}_2$ holds for all $s\in[0,t]$ when $m\geq C_1\bk{\pt{n^{2}\lambda_0^{-1}}^{15}+\pt{\log\mpt{C_2n^2/\delta}}^5}$.
 Finally, by choosing
 \[\poly\mpt{n,\lambda_0^{-1},\log(1/\delta)}=C_1\bk{\pt{n^{2}\lambda_0^{-1}}^{15}+\pt{2n+\log\mpt{1/\delta}+\log C_2}^5},\]
 we can complete the proof of this lemma.
 
\end{proof}

\begin{lemma}
 \label{lem:A_lazy_W}Fix $l \in [L]$, $p \in [2]$ and let $\delta\in(0,1)$, $t\geq 0$. Suppose that
 \[\norm{f_s(\bm{X}) - \bm{y}}_2 \leq \exp\mpt{-\frac{\lambda_0}{4n}s} \norm{\bm{y}}_2\qquad~\text{holds for all}~s\in[0,t],\] then we have the following results: 
\begin{itemize}
 \item $i)$ Suppose that $\norm{ \bm{W}^{(p',l')}_s - \bm{W}^{(p',l')}_0 }_F \leq {\sqrt{m}}/{(\log m)^3}$ holds for all $(p',l')\not= (p,l)$ and $\norm{ \bm{V}^{(p'',l'')}_s - \bm{V}^{(p'',l'')}_0 }_F \leq {\sqrt{m}}/{(\log m)^3}$ holds for all $l''\in[L]$ and $p''\in[2]$ when $s\in[0,t]$. 
 Then there exists a polynomial $\poly(\cdot)$, such that when $m\geq\poly\mpt{n,\norm{\bm{y}}_2,\lambda_0^{-1},\log(1/\delta)}$, 
 with probability at least $1-\delta$, we have
 \begin{align*}
 \sup_{s\in[0,t]}\norm{\bm{W}^{(p,l)}_s-\bm{W}^{(p,l)}_0 }_F = O\mpt{{n\norm{\bm{y}}_2}/{\lambda_0}};
 \end{align*}

 \item $ii)$ Suppose that $\norm{ \bm{V}^{(p',l')}_s - \bm{V}^{(p',l')}_0 }_F \leq {\sqrt{m}}/{(\log m)^3}$ holds for all $(p',l') \not= (p,l)$ and $\norm{ \bm{W}^{(p'',l'')}_s - \bm{W}^{(p'',l'')}_0 }_F \leq {\sqrt{m}}/{(\log m)^3}$ holds for all $l''\in[L]$ and $p''\in[2]$ when $s\in[0,t]$. 
 Then there exists a polynomial $\poly(\cdot)$, such that when $m\geq\poly\mpt{n,\norm{\bm{y}}_2,\lambda_0^{-1},\log(1/\delta)}$, 
 with probability at least $1-\delta$, we have
 \begin{align*}
 \sup_{s\in[0,t]}\norm{\bm{V}^{(p,l)}_t-\bm{V}^{(p,l)}_0 }_F = O\mpt{{n\norm{\bm{y}}_2}/{\lambda_0}}.
 \end{align*}
\end{itemize}

\end{lemma}

\begin{proof}
 First of all, we have
 \begin{align*}
 \norm{\bm{W}^{(p,l)}_{t_0} - \bm{W}^{(p,l)}_0}_F &= \norm{\int^{t_0}_0 \mr{d} \bm{W}^{(p,l)}_s}_F
 =\norm{\int^{t_0}_0 \frac{1}{n}\sum^n_{i=1} ( f^{m}_s(\bm{x}_i) - y_i) \nabla_{\bm{W}^{(p,l)}} f^{m}_s(\bm{x}_i)~\mr{d} s }_F\\
 &\leq \frac{1}{\sqrt{2}n}\sum_{i=1}^n \max_{0\leq s \leq t_0} \norm{\nabla_{\bm{W}^{(p,l)}} f_s^{(p),m}(\bm{x}_i)}_F \int^{t_0}_0 \norm{f^m_s(\bm{X})-\bm{y}}_2~ \mr{d} s\\
 &\leq O\mpt{\frac{\norm{\bm{y}}_2}{\lambda_0}}\cdot \sum_{i=1}^n\max_{0\leq s \leq t_0} \norm{\nabla_{\bm{W}^{(p,l)}} f_s^{(p),m}(\bm{x}_i)}_F
 \end{align*}
for all $t_0\in[0,t]$, and
 \begin{align}\label{eq: sup_Delta_W_lazy} \sup_{t_0\in[0,t]}\norm{\bm{W}^{(p,l)}_{t_0} - \bm{W}^{(p,l)}_0}_F\leq O\mpt{\frac{\norm{\bm{y}}_2}{\lambda_0}}\cdot \sum_{i=1}^n\max_{0\leq s \leq t} \norm{\nabla_{\bm{W}^{(p,l)}} f_s^{(p),m}(\bm{x}_i)}_F.\end{align}
 
 Also, we can get
 \begin{equation*}
 \norm{\nabla_{\bm{W}^{(p,l)}} f_s^{(p),m}(\bm{x}_i) }_F \leq \norm{\nabla_{\bm{W}^{(p,l)}} f_0^{(p),m}(\bm{x}_i) }_F + \norm{\nabla_{\bm{W}^{(p,l)}} f_s^{(p),m}(\bm{x}_i) - \nabla_{\bm{W}^{(p,l)}} f_0^{(p),m}(\bm{x}_i) }_F
 \end{equation*}
 by the triangle inequality. For the first term, by Lemma \ref{lem:NTK_f0}, we know that with probability at least $1-\exp\mpt{-\Omega(m)}$, we have $\norm{\nabla_{ \bm{W}^{(p,l)}} f_0^{(p),m}(\bm{x}_i) }_F = O(1)$ for any $i\in[n]$. So it suffices to bound the second term.

Denote $\mathcal{A} = \left\{s\in[0,t]:\norm{\bm{W}^{(p,l)}_s- \bm{W}^{(p,l)}_0}_F \geq {\sqrt{m}}/{(\log m)^3} \right\}$. Assume that $\mathcal{A}\not=\varnothing$ and let $s_0=\min\mathcal{A}$. Then for any $p'$, $l'$, we have $\norm{\bm{W}^{(p',l')}_s - \bm{W}^{(p',l')}_0}_F \leq {\sqrt{m}}/{(\log m)^3}$ and $\norm{\bm{V}^{(l')}_s - \bm{V}^{(l')}_0}_F \leq {\sqrt{m}}/{(\log m)^3}$ when $s\in[0,s_0]$. 

By Lemma \ref{lem:NTK_ft}, we know for any $i \in [n]$, with probability at least $1-\exp\mpt{-\Omega\big(m(\log m)^{-2}\big)}\geq 1-\exp\mpt{-\Omega(m^{5/6})}$, we have
 \begin{equation}\label{eq: Delta_nabla_W_pl_f_lazy}
 \max_{s\in[0,s_0]}\norm{\nabla_{ \bm{W}^{(p,l)}} f^{(p),m}_s(\bm{x}_i) - \nabla_{ \bm{W}^{(p,l)}} f^{(p),m}_0(\bm{x}_i) }_F = O(1).
 \end{equation}
 Combine with the definition of $s_0$, with probability at least $1-n\exp\mpt{-\Omega(m^{5/6})}$, we have
 \begin{equation*}
{\sqrt{m}}/{(\log m)^3}\leq \norm{\bm{W}^{(p,l)}_{s_0} - \bm{W}^{(p,l)}_0}_F= O\mpt{n\norm{\bm{y}}_2/{\lambda_0}},
 \end{equation*}
which will lead to contradiction when $m\geq \Omega\mpt{n\norm{\bm{y}}_2\lambda_0^{-1}}^5$. This means that $\mathcal A=\varnothing$ and Eq.({\ref{eq: Delta_nabla_W_pl_f_lazy}}) holds for $s_0=t$. Comibine with Eq.(\ref{eq: sup_Delta_W_lazy}), we can get the conclusion of $i)$. Also, it is easy to check that there exists a positive absolute constant $C$ such that when $m
 \geq C\log(n/\delta)^{6/5}$, we have $1-n\exp\mpt{-\Omega(m^{5/6})}\geq 1-\delta$.

Finally, by choosing
 \[\poly\mpt{n,\lambda_0^{-1},\log(1/\delta)}=C'\bk{\pt{n\norm{\bm{y}}_2\lambda_0^{-1}}^{5}+\pt{n+\log\mpt{1/\delta}}^2+1}\]
for some positive absolute constant $C'>0$, we can complete the proof of $i)$. And we can prove $ii)$ with the same above argument.
 
\end{proof}

\begin{lemma}
 \label{lem:A_lazy_regime}There exists a polynomial $\poly(\cdot)$, such that for any $\delta\in(0,1)$, when $m\geq\poly\mpt{n,\norm{\bm{y}}_2,\lambda_0^{-1},\log(1/\delta)}$, then with probability at least $1-\delta$, for all $p\in[2]$ and $l\in[L]$, we have
 $$ \sup_{t\geq 0}\norm{\bm{W}^{(p,l)}_t - \bm{W}^{(p,l)}_0 }_F =O(m^{1/4}), \quad \sup_{t\geq 0}\norm{\bm{V}^{(p,l)}_t - \bm{V}^{(p,l)}_0 }_F =O(m^{1/4}). $$
\end{lemma}

\begin{proof}[Proof of Lemma \ref{lem:A_lazy_regime}]
Denote $t_0 = \min\Big\{t\geq 0:\exists l,~p~\text{such that}~\norm{\bm{W}^{(p,l)}_t - \bm{W}^{(p,l)}_0}_F \geq m^{1/4}~\text{or}~\norm{\bm{V}^{(p,l)}_t - \bm{V}^{(p,l)}_0}_F \geq m^{1/4} ~\text{or}~\norm{\bm{u}(t)}_2 \geq \exp\mbk{-{\lambda_0 t}/{(4n)}} \norm{\bm{y}}_2\Big\} $ and assume that $t_0$ is finite. Then for all $t\in[0,t_0]$, we can get 
\[\norm{\bm{W}^{(p,l)}_t - \bm{W}^{(p,l)}_0}_F \leq m^{1/4}, \quad \norm{\bm{V}^{(p,l)}_t - \bm{V}^{(p,l)}_0 }_F \leq m^{1/4}~\text{and}~\norm{\bm{u}(t)}_2 \leq \exp\mpt{-\frac{\lambda_0 t}{4n}} \norm{\bm{y}}_2\]
hold for all $p$, $l$. According to Lemma \ref{lem:A_lazy_W} and Lemma \ref{lem: smallest eigenvalue leads to fast convergence}, there exists a polynomial $\poly(\cdot)$, such that when $m\geq \poly\mpt{n,\norm{\bm{y}}_2,\lambda_0^{-1},\log(1/\delta)}$, with probability at least $1-\delta$, we have 
\begin{align*}\norm{\bm{W}^{(p,l)}_{t_0}-\bm{W}^{(p,l)}_0 }_F = O\big({{n}\norm{\bm{y}}_2}/{\lambda_0}\big),\qquad~\norm{\bm{V}^{(p,l)}_{t_0}-\bm{V}^{(p,l)}_0 }_F = O\big({{n}\norm{\bm{y}}_2}/{\lambda_0}\big)\end{align*}
hold for all $p$, $l$ and
\[\norm{\bm{u}(t_0)}_2 \leq \exp\mpt{-\frac{\lambda_0 t}{2n}} \norm{\bm{y}}_2.\]
 Combine with the definition of $t_0$, we can get there exist $p$, $l$ such that
\begin{align*}m^{1/4}\leq\norm{\bm{W}^{(p,l)}_{t_0}-\bm{W}^{(p,l)}_0 }_F = O\mpt{\tfrac{{n}\norm{\bm{y}}_2}{\lambda_0}}\quad~\text{or}\quad~m^{1/4}\leq\norm{\bm{V}^{(p,l)}_{t_0}-\bm{V}^{(p,l)}_0 }_F = O\mpt{\tfrac{{n}\norm{\bm{y}}_2}{\lambda_0}}.\end{align*}
However, this will lead to contradiction when $m\geq C\mpt{n\norm{\bm{y}}_2\lambda_0^{-1}}^5$ for some positive absolute constant $C>0$.
\end{proof}

\subsection{Nearly Hölder Continuity of $r_{\theta(t)}^{m}$}


\begin{lemma}\label{lem:8.2_x_z} 
Let $\tau\in\bk{\Omega\mpt{1/\sqrt{m}},O\mpt{{\sqrt{m}}/{(\log m)^3}}}$, $T\subseteq [0,\infty)$ and fix $\bm{z}\in\mb{S}^{d-1}$. Suppose that $\norm{\bm{W}^{(l)}_t - \bm{W}^{(l)}_0 }_{F}\leq\tau$ and $\norm{\bm{V}^{(l)}_t - \bm{V}^{(l)}_0 }_{F}\leq\tau$ hold for all $t\in T$ and $l\in[L]$. Then there exists a positive absolute constant $C$, such that with probability at least $1-\exp\mpt{-\Omega(m^{2/3}\tau^{2/3})}$, for all $t\in T$, $l\in[L]$ and $\bm{x}\in\mb{S}^{d-1}$ such that $\norm{\bm{x}-\bm{z}}_2\leq O\mpt{1/m}$, we have 
\begin{itemize}
 \item $i)$ $\norm{\bm{g}'_{l,\bm{xz}}}_2 = O\mpt{{\tau}/{\sqrt{m}}}$;
 \item $ii)$ $\norm{\bm{D}^{(l)'}_{\bm{xz}}}_0 =O\mpt{m^{2/3}\tau^{2/3}}$ and $\norm{\bm{D}^{(l)'}_{\bm{xz}} \bm{W}^{(l)}_t \bm{\alpha}^{(l-1)}_{t,\bm{x}}}_2=O\mpt{{\tau}}$;
 \item $iii)$ $\norm{\Delta\bm{\alpha}^{(l)}_{\bm{xz}}}_2=O\mpt{{\tau}/{\sqrt{m}}}$.
\end{itemize}
when $m$ is greater than the positive constant $C$. 
\end{lemma}

\begin{proof}[Proof of Lemma \ref{lem:8.2_x_z}]The proof of this lemma is similar to the proof of Lemma \ref{lem: 3result_point}. The only thing to note is that the conclusion of this lemma holds uniformly for $\bm{x}\in\mb{S}^{d-1}$ such that $\norm{\bm{x}-\bm{z}}_2\leq O\mpt{1/m}$ with high probability for any fixed $\bm{z}$. We inductively prove this lemma. 

Since $\bm{A}$ does not change during the training, we can easily check that, with probability at least $1-\exp\mpt{-\Omega(m)}$, we have $\norm{\Delta\bm{\alpha}^{(0)}_{\bm{xz}}}_2=\norm{\bm{A}(\bm{x}-\bm{z})/\sqrt{m}}_2=O\mpt{1/{{m}}}$ for any $\bm{x}\in\mb{S}^{d-1}$ such that $\norm{\bm{x}-\bm{z}}_2\leq O(1/m)$, which means that $iii)$ holds for $l=0$ since $\tau=\Omega(1/\sqrt{m})$. Then it only needs to be proven that
\[iii)~\text{holds for}~l=k\quad\Longrightarrow\quad~\text{Lemma holds for}~l=k+1.\]
Now we assume that $iii)$ holds for $l=k\in\{0,1,\cdots,L-1\}$, then with probability at least $1-\exp\mpt{-\Omega(m)}$, for any $\bm{x}\in\mb{S}^{d-1}$ such that $\norm{\bm{x}-\bm{z}}_2=O(1/m)$, we have
\[\norm{\bm{\alpha}^{(k)}_{t,\bm{x}}}_2=\norm{\bm{\alpha}^{(k)}_{0,\bm{z}}+\Delta\bm{\alpha}^{(k)}_{\bm{xz}}}_2\leq\norm{\bm{\alpha}^{(k)}_{0,\bm{z}}}_2+\norm{\Delta\bm{\alpha}^{(k)}_{\bm{xz}}}_2=O(1)+O\mpt{\tau/\sqrt{m}}=O(1).\]

For $i)$, similar to the proof of Lemma \ref{lem: 3result_point}, we can get
\begin{align*}
 \bm{g}_{k+1,\bm{xz}}'&=\sqrt{\frac{2}{m}}\pt{ \Delta\bm{W}^{(k+1)} \bm{\alpha}^{(k)}_{t,\bm{x}} + \bm{W}^{(k+1)}_0 \Delta\bm{\alpha}^{(k)}_{\bm{xz}}}. 
 \end{align*}
Thus we can get $i)$ holds for $l=k+1$.

Considering that the conclusion of Lemma \ref{lem: D'_Estimation} holds uniformly for $\bm{g}'$ with high probability, taking $\bm{W}_0=\bm{W}^{(k+1)}_0$, $\bm{h}=\bm{\alpha}^{(k)}_{0,\bm{z}}$, $\bm{g}'=\bm{g}'_{k+1,\bm{xz}}$ and $\delta=\Theta\mpt{{\tau}/{\sqrt{m}}}\leq O\mpt{(\log m)^{-3}}$, then we can get $ii)$ holds for $l=k+1$ with probability at least $1-\exp\mpt{-\Omega(m^{2/3}\tau^{2/3})}$.

As for $iii)$, it is easy to check that
\begin{align*}
\Delta\bm{\alpha}^{(k+1)}_{\bm{{xz}}}
&=\Delta\bm{\alpha}^{(k)}_{\bm{xz}}+\frac{\sqrt{2}a}{m}\Big[\bm{V}^{(k+1)}_t\bm{D}^{(k+1)'}_{\bm{xz}}\bm{W}^{(k+1)}_t\bm{\alpha}^{(k)}_{t,\bm{x}}+\Delta\bm{V}^{(k+1)}\bm{D}^{(k+1)}_{0,\bm{z}}\bm{W}^{(k+1)}_t\bm{\alpha}^{(k)}_{t,\bm{x}}
\\&\qquad\qquad\qquad\qquad\qquad\qquad+\bm{V}^{(k+1)}_0\bm{D}^{(k+1)}_{0,\bm{z}}\pt{\bm{W}^{(k+1)}_t\bm{\alpha}^{(k)}_{t,\bm{x}}-\bm{W}^{(k+1)}_0\bm{\alpha}^{(k)}_{0,\bm{z}}}\Big].
\end{align*}
We have shown that, with probability at least $1-\exp\mpt{-\Omega(m^{2/3}\tau^{2/3})}$, $i)$ $ii)$ hold for $l=k+1$. Combine with Lemma \ref{lem:up_bound_of_alpha_delta}, we can get $\norm{\Delta\bm{\alpha}^{(k+1)}_{\bm{xz}}}_2= O\mpt{\tau/\sqrt{m}}$. 

 Thus, we finish the proof.
 
\end{proof}

\begin{lemma}\label{lem:8.2_Delta_delta_xz}
Let $\tau\in\bk{\Omega\mpt{1/\sqrt{m}},O\mpt{{\sqrt{m}}/{(\log m)^3}}}$, $T\subseteq [0,\infty)$ and fix $\bm{z}\in\mb{S}^{d-1}$. Suppose that $\norm{\bm{W}^{(l)}_t - \bm{W}^{(l)}_0 }_{F}\leq\tau$ and $\norm{\bm{V}^{(l)}_t - \bm{V}^{(l)}_0 }_{F}\leq\tau$ hold for all $t\in T$ and $l\in[L]$. Then there exists a positive absolute constant $C$, such that with probability at least $1-\exp\mpt{-\Omega(m^{2/3}\tau^{2/3})}$, for all $t\in T$, $l\in[L]$ and $\bm{x}\in\mb{S}^{d-1}$ such that $\norm{\bm{x}-\bm{z}}_2\leq O\mpt{1/m}$, we have 
\begin{align*}
\norm{\Delta\bm{\delta}^{(l)}_{\bm x \bm z}}_2= O\mpt{m^{1/3}\tau^{1/3}\sqrt{\log m}},
\end{align*}
when $m$ is greater than the positive constant $C$. 
\end{lemma}

\begin{proof}[Proof of Lemma \ref{lem:8.2_Delta_delta_xz}]
The proof of this lemma is similar to the proof of Lemma \ref{lem: Delta_delta_point}. The only thing to note is that the conclusion of this lemma holds uniformly for $\bm{x}\in\mb{S}^{d-1}$ such that $\norm{\bm{x}-\bm{z}}_2\leq O\mpt{1/m}$ with high probability for any fixed $\bm{z}$. We inductively prove this lemma. 

Base case: $\norm{\Delta\bm{\delta}^{(L)}_{\bm{xz}}}_2 = 0$ since $\bm{v}$ is fixed during the training process.

Assume that this lemma holds for $l+1$, then with probability at least $1-\exp\mpt{-\Omega(m)}$, we have
\[\norm{\bm{\delta}^{(l+1)}_{t,\bm{x}}}_2=\norm{\Delta\bm{\delta}^{(l+1)}_{\bm{xz}}+\bm{\delta}^{(l+1)}_{0,\bm{z}}}_2\leq\norm{\Delta\bm{\delta}^{(l+1)}_{\bm{xz}}}_2+\norm{\bm{\delta}^{(l+1)}_{0,\bm{z}}}_2\leq O(\sqrt{ m})\]
for all $\bm{x}\in\mb{S}^{d-1}$. Moreover, it is easy to check that
\begin{align*}
\Delta\bm{\delta}^{(l)}_{\bm{xz}}
&=\Delta\bm{\delta}^{(l+1)}_{\bm{xz}}+\tfrac{\sqrt{2}a}{m}\Big[\bm{W}^{(l+1),T}_{0}\bm{D}^{(l+1)'}_{\bm{xz}}\bm{V}^{(l+1),T}_{0}\bm{\delta}^{(l+1)}_{0,\bm{z}}+\Delta\bm{W}^{(l+1),T}\bm{D}^{(l+1)}_{t,\bm{x}}\bm{V}^{(l+1),T}_0\bm{\delta}^{(l+1)}_{0,\bm{x}}
\\&\qquad\qquad\qquad+\bm{W}^{(l+1),T}_t\bm{D}^{(l+1)}_{t,\bm{x}}\Delta\bm{V}^{(l+1),T} \bm{\delta}^{(l+1)}_{0,\bm{x}} + \bm{W}^{(l+1),T}_t\bm{D}^{(l+1)}_{t,\bm{x}}\bm{V}^{(l+1),T}_t \Delta\bm{\delta}^{(l+1)}_{\bm{xz}}\Big].
\end{align*}
Let us denote the four terms within the square brackets, excluding the factor `$\sqrt 2a/m$' outside the brackets, as $\bm{u}_1$ to $\bm{u}_4$ respectively. We can control $\norm{\bm{u}_2}_2$, $\norm{\bm{u}_3}_2$, and $\norm{\bm{u}_4}_2$ using the same method as in the proof of Lemma \ref{lem: Delta_delta_point}, since $\norm{\bm{D}^{(l+1)}_{t,\bm{x}}}_2\leq 1$ and the conclusion of Lemma \ref{lem:up_bound_of_alpha_delta} holds uniformly for $\bm{x}$.

As for $\bm{u}_1$, if ${\bm{\delta}^{(l+1)}_{0,\bm{z}}}=\bm{0}$ or $\norm{\bm{D}^{(l+1)'}_{\bm{xz}}}_{0}=0$, we have $\norm{\bm{u}_1}_2=0$. Therefore, we consider the case where ${\bm{\delta}^{(l+1)}_{0,\bm{z}}}\not=\bm{0}$ and $\norm{\bm{D}^{(l+1)'}_{\bm{xz}}}_0\geq1$. Denote $\tilde{\bm{\delta}}_{\bm{z}}={\bm{\delta}^{(l+1)}_{0,\bm{z}}}/\norm{\bm{\delta}^{(l+1)}_{0,\bm{z}}}_2$ for $\bm{\delta}^{(l+1)}_{0,\bm{z}}\in\mb{R}^{m}\backslash\{\bm{0}\}$, we can get
\begin{align*}
\norm{\bm{u}_1}_2&\leq \norm{\bm{W}^{(l+1)}_0}_2\norm{\bm{D}^{(l+1)'}_{\bm{xz}}\bm{V}^{(l+1)}_0\tilde{\bm{\delta}}_{\bm{z}}}_2\norm{\bm{\delta}^{(l+1)}_{0,\bm{z}}}_2\leq O\mpt{m}\norm{\bm{D}^{(l+1)'}_{\bm{xz}}\bm{V}^{(l+1)}_0\tilde{\bm{\delta}}_{\bm{z}}}_2.
\end{align*}
Using the randomness of $\bm{V}^{(l+1)}_0$, for any fixed $\tilde{\bm{\delta}}_{\bm{z}}$, we have $\bm{V}^{(l+1)}_0\tilde{\bm{\delta}}_{\bm{z}}\sim\mathcal{N}(\bm{0},\bm{I}_m)$. Thus, by Lemma \ref{lem:Recur_uv} and taking $s\geq\norm{\bm{D}^{(l+1)'}_{\bm{xz}}}_0$, with probability at least $1-\exp\mpt{-\Omega(s\log m)}$, we can get
\begin{equation*}
 \forall \bm{u} \in \mathbb{R}^m\, \text{s.t.} \norm{\bm{u}}_0 \leq s, \quad\text{we have }\quad \abs{ \bm{u}^T \bm{V}^{(l+1)}_0\tilde{\bm{\delta}}_{\bm{z}}} \leq 9 \sqrt{s \log m } \norm{\bm{u}}_2.
 \end{equation*}
According to Lemma \ref{lem:8.2_x_z}, with probability at least $1-\exp\mpt{-\Omega(m^{2/3}\tau^{2/3})}$, for any $\bm{x}\in\mb{S}^{d-1}$ such that $\norm{\bm{x}-\bm{z}}_2\leq O\mpt{1/m}$, we have $\norm{\bm{D}^{(l)'}_{\bm{xz}}}_0 =O\mpt{m^{2/3}\tau^{2/3}}$. By taking $s=\Theta(m^{2/3}\tau^{2/3})$, we can get
\[\norm{\bm{D}^{(l+1)'}_{\bm{xz}}\bm{V}^{(l+1)}_0\tilde{\bm{\delta}}_{\bm{z}}}_2=\sup_{\bm{u}\in\mb{S}^{d-1}}\abs{ \bm{u}^T\bm{D}^{(l+1)'}_{\bm{xz}}\bm{V}^{(l+1)}_0\tilde{\bm{\delta}}_{\bm{z}}} \leq 9 \sqrt{s \log m }\]
holds uniformly for $\bm{x}$.

Combining the above discussions, we can conclude that $\norm{\Delta\bm{\delta}^{(l)}_{\bm x \bm z}}_2\leq O\mpt{m^{1/3}\tau^{1/3}\sqrt{\log m}}$.

\end{proof}

\begin{lemma}\label{lem: gamma_eta_t_vs_0_uniform}Let $\tau\in\bk{\Omega\mpt{1/\sqrt{m}},O\mpt{{\sqrt{m}}/{(\log m)^3}}}$, $T\subseteq [0,\infty)$ and fix $l\in[L]$, $\bm{z}\in\mb{S}^{d-1}$. Suppose that $\norm{\bm{W}^{(l)}_t - \bm{W}^{(l)}_0 }_{F}\leq\tau$ and $\norm{\bm{V}^{(l)}_t - \bm{V}^{(l)}_0 }_{F}\leq\tau$ hold for all $t\in T$, then there exists a positive absolute constant $C$, such that with probability at least $1-\exp\mpt{-\Omega(m^{2/3}\tau^{2/3})}$ over the randomness of $\bm{W}^{(l)}_0$ and $\bm{V}^{(l)}_0$, for all $t\in T$, $l\in[L]$ and $\bm{x}\in\mb{S}^{d-1}$ such that $\norm{\bm{x}-\bm{z}}_2\leq O\mpt{1/m}$, we have
 \begin{align*}
 \begin{gathered}
 \norm{\Delta\bm{\gamma}^{(l)}_{\bm x \bm z}}_{2}= O\mpt{m^{-1/6}\tau^{1/3}\sqrt{\log m}}, \quad~ \norm{\Delta\bm{\eta}^{(l)}_{\bm x \bm z}}_{2} = O\mpt{\frac{\tau}{{m}}};\\
 \norm{\bm{\gamma}^{(l)}_{t,\bm x}}_2=O(1),\qquad~ \norm{\bm{\eta}^{(l)}_{t,\bm x}}_2 =O(1/\sqrt{m})
 \end{gathered}
 \end{align*}
when $m$ is greater than the positive constant $C$. 
\end{lemma}

\begin{proof}[Proof of Lemma \ref{lem: gamma_eta_t_vs_0_uniform}]
First of all, we have
\begin{align*}
&\bm{\gamma}^{(l)}_{t,\bm x} -\bm{\gamma}^{(l)}_{0,\bm z}=\frac{\sqrt{2}}{m}\pt{\bm{D}^{(l)}_{t,\bm x} \bm{V}^{(l),T}_t 
 \bm{\delta}^{(l)}_{t,\bm x}- \bm{D}^{(l)}_{0,\bm z} \bm{V}^{(l),T}_0 \bm{\delta}^{(l)}_{0,\bm z}}\\
&\qquad=\frac{\sqrt{2}}{m}\pt{\bm{D}^{(l)'}_{\bm x \bm z}\bm{V}^{(l),T}_0\bm{\delta}^{(l)}_{0,\bm z}+\bm{D}^{(l)}_{t,\bm x}\Delta\bm{V}^{(l),T}\bm{\delta}^{(l)}_{t,\bm x}+\bm{D}^{(l)}_t\bm{V}^{(l),T}_0\Delta\bm{\delta}^{(l)}_{\bm x \bm z}}.
\end{align*}
Using the similar proof technique as the previous lemma, we can establish that with probability at least $1-\exp\mpt{-\Omega(m^{2/3}\tau^{2/3})}$, we have
\[\norm{\bm{D}^{(l)'}_{\bm x \bm z}\bm{V}^{(l),T}_0\bm{\delta}^{(l)}_{0,\bm z}}_2=O\mpt{{m^{5/6}\tau^{1/3}\sqrt{\log m}}}.\]
According to Corollary \ref{coro: Random matrix}, Lemma \ref{lem:up_bound_of_alpha_delta} and Lemma \ref{lem:8.2_Delta_delta_xz} we can get
\begin{align*}
\norm{\bm{D}^{(l)}_{t,\bm x}\Delta\bm{V}^{(l),T}\bm{\delta}^{(l)}_{t,\bm x}}_2=O\mpt{\tau\sqrt{m}};\qquad~\norm{\bm{D}^{(l)}_t\bm{V}^{(l),T}_0\Delta\bm{\delta}^{(l)}_{\bm x \bm z}}_2=O\mpt{m^{5/6}\tau^{1/3}\log m}.
\end{align*}
Thus, we can get $\norm{\Delta\bm{\gamma}^{(l)}_{\bm x \bm z}}_{2}= O\mpt{m^{-1/6}\tau^{1/3}\log m}$.

As for $\norm{\Delta\bm{\eta}^{(l)}_{\bm x \bm z}}_{2}$, we can similarly get
\begin{align*}
\norm{\Delta\bm{\eta}^{(l)}_{\bm x \bm z}}_2&=\frac{\sqrt{2}}{m}\norm{\bm{D}^{(l)'}_{\bm x \bm z}\bm{W}^{(l)}_t\bm{\alpha}^{(l-1)}_{t,x}+\bm{D}^{(l)}_{0,\bm z}\Delta\bm{W}^{(l)}\bm{\alpha}^{(l-1)}_{t,\bm x}+\bm{D}^{(l)}_{0,\bm z}\bm{W}^{(l)}_0\Delta\bm{\alpha}^{(l-1)}_{\bm x \bm z}}_2\\
&\leq \frac{\sqrt{2}}{m}\big(O\mpt{\tau}+O\mpt{\tau}+O\mpt{\tau}\big)=O\mpt{\frac{\tau}{m}}
\end{align*}
according to Corollary \ref{coro: Random matrix}, Lemma \ref{lem:up_bound_of_alpha_delta} and Lemma \ref{lem:8.2_x_z}.
With the above results, we can easily get 
\begin{align*}
\begin{gathered}
\norm{\bm{\gamma}^{(l)}_{t,\bm x}}_2\leq\norm{\bm{\gamma}^{(l)}_{0,\bm z}}_2+\norm{\Delta\bm{\gamma}^{(l)}_{\bm x \bm z}}_2=O(1),\quad\norm{\bm{\eta}^{(l)}_{t,\bm x}}_2\leq\norm{\bm{\eta}^{(l)}_0}_2+\norm{\Delta\bm{\eta}^{(l)}_{\bm x \bm z}}_2=O\mpt{\tfrac{1}{\sqrt{m}}}
 \end{gathered}
\end{align*}
since $\tau=O\mpt{{\sqrt{m}}/{(\log m)^3}}$.
\end{proof}

\begin{lemma}
 \label{lem:NTK_ft_uniform}
Let $\tau\in\bk{\Omega\mpt{1/\sqrt{m}},O\mpt{{\sqrt{m}}/{(\log m)^3}}}$, $T\subseteq [0,\infty)$ and fix $\bm{z}\in\mb{S}^{d-1}$. Suppose that $\norm{\bm{W}^{(l)}_t - \bm{W}^{(l)}_0 }_{F}\leq\tau$ and $\norm{\bm{V}^{(l)}_t - \bm{V}^{(l)}_0 }_{F}\leq\tau$ hold for all $t\in T$ and $l\in[L]$. Then there exists a positive absolute constant $C$, such that with probability at least $1-\exp\mpt{-\Omega(m^{2/3}\tau^{2/3})}$, for all $l\in[L]$ and $\bm{x}\in\mb{S}^{d-1}$ such that $\norm{\bm{x}-\bm{z}}_2\leq O\mpt{1/m}$, we have 
\begin{align*}
\begin{gathered}
 \sup_{t\in T}\norm{\nabla_{\bm{W}^{(l)}} f^{(p),m}_t(\bm{x}) - \nabla_{\bm{W}^{(l)}} f^{(p),m}_0(\bm{z})}_{F}= O\mpt{m^{-1/6}\tau^{1/3} \sqrt{\log m} };\\
 \sup_{t\in T}\norm{\nabla_{\bm{V}^{(l)}} f^{(p),m}_t(\bm{x}) - \nabla_{\bm{V}^{(l)}} f^{(p),m}_0(\bm{z})}_{F}= O\mpt{m^{-1/6}\tau^{1/3}\sqrt{\log m}},
 \end{gathered}
 \end{align*}
 when $m$ is greater than the positive constant $C$. 
\end{lemma}

\begin{proof}[Proof of Lemma \ref{lem:NTK_ft_uniform}]According to Eq.(\ref{eq: nabla_W_V_f}), we have
\begin{align*}
 &\norm{\nabla_{\bm{W}^{(l)}} f^{(p),m}_t(\bm{x})- \nabla_{\bm{W}^{(l)}} f^{(p),m}_0(\bm{z})}_{F} = \norm{ a \bm{\gamma}^{(l)}_{t,\bm{x}} \bm{\alpha}^{(l-1),T}_{t,\bm{x}} - a \bm{\gamma}^{(l)}_{0,\bm z}\bm{\alpha}^{(l-1),T}_{0,\bm z}}_{F}\\
 &\qquad=a\norm{\bm{\gamma}^{(l)}_{t,\bm x}\Delta\bm{\alpha}^{(l-1),T}_{\bm x \bm z}+\Delta\bm{\gamma}^{(l)}_{\bm x \bm z}\bm{\alpha}^{(l-1),T}_{0,\bm z}}_F\leq a\norm{\bm{\gamma}^{(l)}_{t,\bm x}\Delta\bm{\alpha}^{(l-1),T}_{\bm x \bm z}}_F+a\norm{\Delta\bm{\gamma}^{(l)}_{\bm x \bm z}\bm{\alpha}^{(l-1),T}_{0,\bm z}}_F\\
 &\qquad= a\norm{\bm{\gamma}^{(l)}_{t,\bm x}}_2 \norm{\Delta\bm{\alpha}^{(l-1)}_{\bm x\bm z}}_{2} + a\norm{\Delta\bm{\gamma}^{(l)}_{\bm x \bm z}}_2\norm{\bm{\alpha}^{(l-1)}_{0,\bm z}}_2 \leq O\mpt{m^{-1/6}\tau^{1/3} \sqrt{\log m}}
 \end{align*}
according to Lemmas \ref{lem: gamma_eta_t_vs_0_uniform} and \ref{lem:8.2_x_z} $iii)$. Similarly, we can also get
 \begin{align*}
 &\norm{\nabla_{\bm{V}^{(l)}} f^{(p),m}_t(\bm{z}) - \nabla_{\bm{V}^{(l)}} f^{(p),m}_0(\bm{z})}_{F} = \norm{a \bm{\delta}^{(l)}_{t,\bm x} \bm{\eta}^{(l),T}_{t,\bm x} - a \bm{\delta}^{(l)}_{0,\bm z} \bm{\eta}^{(l),T}_{0,\bm z}}_{F}\\
 &\qquad\leq a\norm{\bm{\eta}^{(l)}_{t,\bm x}}_2 \norm{\Delta\bm{\delta}^{(l)}_{\bm x\bm z}}_{2} + a\norm{\Delta\bm{\eta}^{(l)}_{\bm x \bm z}}_2 \norm{\bm{\delta}^{(l)}_{0,\bm z}}_2\leq O\mpt{m^{-1/6}\tau^{1/3} \sqrt{\log m}}
 \end{align*}
according to Lemmas \ref{lem: gamma_eta_t_vs_0_uniform} and \ref{lem:8.2_Delta_delta_xz}.

Thus, we finish the proof.

\end{proof}

\begin{proposition}
 \label{prop:During training_uniform}Fix $ \bm{z},\bm{z}' \in \mathbb{S}^{d-1}$ and let $\delta\in(0,1)$, $T\subseteq[0,\infty)$. Suppose that $\norm{\bm{W}^{(l)}_t - \bm{W}^{(l)}_0}_{F}= O(m^{1/4})$ and $\norm{\bm{V}^{(l)}_t - \bm{V}^{(l)}_0 }_{F}= O(m^{1/4})$ hold for all $l\in [L]$ and $t\in T$. Then there exist some positive absolute constants $C_1>0$ and $C_2\geq 1$, such that with probability at least $1-\delta$, for any $ \bm{x},\bm{x}' \in \mathbb{S}^{d-1}$ such that $\norm{\bm{x}- \bm{z}}_2,\norm{\bm{x}'-\bm{z}'}_2 \leq O(1/m)$, we have
 \[\sup_{t\in T}\abs{r_t^{(p),m}(\bm{x},\bm{x}') - r_0^{(p),m}(\bm{z},\bm{z}')}= O \mpt{m^{-\frac{1}{12}}\sqrt{\log m}},~\text{when}~m\geq C_1\pt{\log(C_2/\delta)}^{6/5}.\]
\end{proposition}
\begin{proof}[Proof of Proposition \ref{prop:During training_uniform}]
By Lemma \ref{lem:NTK_ft_uniform} (choose parameter $\tau = \Theta(m^{1/4})$), Lemma \ref{lem:NTK_f0} and
\begin{align*}
 \norm{ \nabla_{\bm{W}^{(l)}} f^{(p),m}_t(\bm{x})}_{F}&\leq \norm{ \nabla_{\bm{W}^{(l)}} f^{(p),m}_0(\bm{z})}_{F}+\norm{ \nabla_{\bm{W}^{(l)}} f^{(p),m}_t(\bm{x})- \nabla_{\bm{W}^{(l)}} f^{(p),m}_0(\bm{z}) }_{F};\\
 \norm{ \nabla_{\bm{V}^{(l)}} f^{(p),m}_t(\bm{x}')}_{F} &\leq \norm{ \nabla_{\bm{V}^{(l)}} f^{(p),m}_0(\bm{z}')}_{F}+\norm{ \nabla_{\bm{V}^{(l)}} f^{(p),m}_t(\bm{x}')- \nabla_{\bm{V}^{(l)}} f^{(p),m}_0(\bm{z}') }_{F},
 \end{align*}
with probability at least $1-\exp(-\Omega(m^{5/6}))$, we have
\begin{align*}
&\abs{\ag{\nabla_{\bm{W}^{(l)}} f^{(p),m}_{t}(\bm{x}) , \nabla_{\bm{W}^{(l)}} f^{(p),m}_{t}(\bm{x}')}-\ag{\nabla_{\bm{W}^{(l)}} f^{(p),m}_{0}(\bm{z}) , \nabla_{\bm{W}^{(l)}} f^{(p),m}_{0}(\bm{z}')}}\\
&\qquad\qquad\qquad~~\leq \norm{ \nabla_{\bm{W}^{(l)}} f^{(p),m}_0(\bm{z})}_{F} \norm{ \nabla_{\bm{W}^{(l)}} f^{(p),m}_t(\bm{x}') - \nabla_{\bm{W}^{(l)}} f^{(p),m}_0(\bm{z}')}_{F}\\
&\qquad\qquad\qquad\qquad\qquad\qquad~~+\norm{ \nabla_{\bm{W}^{(l)}} f^{(p),m}_t(\bm{x}')}_{F} \norm{ \nabla_{\bm{W}^{(l)}} f^{(p),m}_t(\bm{x})- \nabla_{\bm{W}^{(l)}} f^{(p),m}_0(\bm{z}) }_{F}\\
&\qquad\qquad\qquad~~ \leq O(1)\cdot O\mpt{m^{-\frac{1}{12}}\sqrt{\log m}}+O(1)\cdot O\mpt{m^{-\frac{1}{12}}\sqrt{\log m}}\leq O\mpt{m^{-\frac{1}{12}}\sqrt{\log m}}
\end{align*}
and similarly have
\begin{align*}\abs{\ag{\nabla_{\bm{V}^{(l)}} f^{(p),m}_{t}(\bm{x}) , \nabla_{\bm{V}^{(l)}} f^{(p),m}_{t}(\bm{x}')}-\ag{\nabla_{\bm{V}^{(l)}} f^{(p),m}_{0}(\bm{z}) , \nabla_{\bm{V}^{(l)}} f^{(p),m}_{0}(\bm{z}')}}\\\leq O\mpt{m^{-\frac{1}{12}}\sqrt{\log m}}
\end{align*}
for all $l\in[L]$, $t\in T$ and $ \bm{x},\bm{x}' \in \mathbb{S}^{d-1}$ such that $\norm{\bm{x}- \bm{z}}_2,\norm{\bm{x}'-\bm{z}'}_2 \leq O(1/m)$ when $m$ is greater than some positive absolute constant $C$. Combine with Eq.(\ref{eq: RNK_formulate}), with probability at least $1-\exp(-\Omega(m^{5/6}))$, we can get
\begin{align*}\sup_{t\in T}\abs{r_t^{(p),m}(\bm{x},\bm{x}') - r_0^{(p),m}(\bm{z},\bm{z}')}= O \mpt{m^{-\frac{1}{12}}\sqrt{\log m}}.
\end{align*}
Also, it is easy to check that there exist some positive absolute constants $C_1>0$ and $C_2\geq 1$ such that $C_1\pt{\log\mpt{C_2/\delta}}^{6/5}\geq C$ holds for $\delta\in(0,1)$ and when $m\geq C_1\pt{\log\mpt{C_2/\delta}}^{6/5}$, we have $1-\exp\mpt{-\Omega\mpt{m^{5/6}}} \geq 1-\delta$.

\end{proof}

\subsection{Hölder Continuity of $r$}

For our convenience let us first introduce the following definition of Hölder spaces.
For a compact set $\Omega$ and $\alpha \in [0,1]$, let us define a semi-norm for $f : \Omega \to \mathbb{R}$ by
\begin{align*}
 \abs{f}_{0,\alpha} = \sup_{x,y \in \Omega,~x\neq y}\frac{\abs{f(x) - f(y)}}{\norm{x-y}^\alpha}
\end{align*}
and define the Hölder space by
\begin{align}
 C^{0,\alpha}(\Omega) = \left\{ f \in C(\Omega) : \abs{f}_{0,\alpha} < \infty \right\},
\end{align}
which is equipped with norm $\norm{f}_{C^{0,\alpha}(\Omega)} = \sup_{x \in \Omega} \abs{f(x)} + \abs{f}_{\alpha}$.
Then it is easy to show that
\begin{itemize}
 \item $i)$ $C^{0,\alpha}(\Omega) \subseteq C^{0,\beta}(\Omega)$ if $\beta \leq \alpha$;
 \item $ii)$ if $f,g \in C^{0,\alpha}(\Omega)$, then $f + g,~ fg \in C^{\alpha}(\Omega)$;
 \item $iii)$ if $f \in C^{0,\alpha}(\Omega_1)$ and $g \in C^{0,\beta}(\Omega_2)$ with $\text{Ran } g \subseteq \Omega_1$, then $f\circ g \in C^{0,\alpha\beta}(\Omega_2)$.
\end{itemize}

\begin{proposition}\label{prop:continuity of NTK}
 We have $r \in C^{0,s}(\Omega)$ with $s = 2^{-L}$ and $\Omega = \mathbb{S}^{d-1} \times \mathbb{S}^{d-1}$, that is,
 there is some constant $C>0$ that
\begin{align*}
 |r(\bm{x},\bm{x}')-r(\bm{z},\bm{z}')| \leq C\|(\bm{x},\bm{x}') - (\bm{z},\bm{z}') \|_2^s. 
\end{align*}
\end{proposition}

\begin{proof}[Proof of Proposition \ref{prop:continuity of NTK}]
Recall that $r$ is given by
\begin{align*}
 r(\bm{x},\bm{x}') = C\sum_{l=1}^L B_{l+1}(\bm{x},\bm{x}')\left[(1+a^2)^{l-1}\kappa_1\left(\frac{K_{l-1}(\bm{x},\bm{x}')}{(1+a^2)^{l-1}}\right)+K_{l-1}(\bm{x},\bm{x}')\kappa_0\left(\frac{K_{l-1}(\bm{x},\bm{x}')}{(1+a^2)^{l-1}}\right)\right],
\end{align*}
where $C={1}/{2L(1+a^2)^{L-1}}$, $K_{0}(\bm{x},\bm{x}') =\bm{x}^{\top}\bm{x}'$, $B_{L+1}(\bm{x},\bm{x}')=1$ and
\begin{gather*}
 \kappa_0(u) = \frac{1}{\pi}(\pi-\arccos(u), \quad \kappa_1(u) = \frac{1}{\pi}\left(u(\pi-\arccos(u))+\sqrt{1-u^2}\right)\\
 K_{l}(\bm{x},\bm{x}') =K_{l-1}(\bm{x},\bm{x}') + a^2 (1+a^2)^{l-1}\kappa_1\left(\frac{K_{l-1}(\bm{x},\bm{x}')}{(1+a^2)^{l-1}}\right)\\
 B_l(\bm{x},\bm{x}') = B_{l+1}(\bm{x},\bm{x}')\left[1+ a^2\kappa_0\left(\frac{K_{l-1}(\bm{x},\bm{x}')}{(1+a^2)^{l-1}}\right)\right].
\end{gather*}

Since $r$ is symmetric, by triangle inequality it suffices to prove that $r(\bm x_0,\cdot) \in C^{0,s}(\mathbb{S}^{d-1})$
 with $\abs{r(\bm x_0,\cdot)}_{0,s}$ bounded by a constant independent of $\bm x_0$.
It is easy to check that $\bm x\to \bm x^T \bm x_0 \in C^{0,1}(\mathbb{S}^{d-1})$
and
\[|\arccos \mu-\arccos \nu|=O(\sqrt{|\mu-\nu|})~\text{and}~|\sqrt{1-\mu^2}-\sqrt{1-\nu^2}|=O(\sqrt{|\mu-\nu|}),\] 
meaning that $\kappa_0,\kappa_1 \in C^{0,\frac{1}{2}}([-1,1])$. Thus, $r \in C^{0,s}(\Omega)$ with $s = (1/2)^{L}$.

\end{proof}

\subsection{Proof of the kernel uniform convergence}

\begin{proof}[Proof of Proposition 3.2]

By Lemma \ref{lem:A_lazy_regime}, there exists a polynomial $\poly_1(\cdot)$, such that for any $\delta\in(0,1)$, when $m\geq\poly_1\mpt{n,\norm{\bm{y}}_2,\lambda_0^{-1},\log(1/\delta)}$, then with probability at least $1-\delta/2$, for all $p\in[2]$ and $l\in[L]$, we have
 $$ \sup_{t\geq 0}\norm{\bm{W}^{(p,l)}_t - \bm{W}^{(p,l)}_0 }_F =O(m^{1/4}), \quad \sup_{t\geq 0}\norm{\bm{V}^{(p,l)}_t - \bm{V}^{(p,l)}_0 }_F =O(m^{1/4}). $$


Since $\|\bm{x}\|_2=1$, we have an $\varepsilon$-net $\mathcal{N}_\varepsilon$ of $\mathcal{X}$ such that the cardinality $|\mathcal{N}_{\varepsilon}|=O(\varepsilon^{-(d-1)})$. We choose $\varepsilon={1}/{m^{2^L}}$ and thus $\log |\mathcal{N}_{\varepsilon}| = O(\log m)$. Denote $B_{\bm{z}}(\varepsilon)=\{\bm{x}\in \mathbb{S}^{d-1}:\|\bm{x}-\bm{z}\|_2\leq\varepsilon\}$. Then, fixing $\bm{z},\bm{z}'\in\mathcal{N}_\varepsilon$, for any $\bm{x} \in B_{\bm z}(\varepsilon)$ and $\bm{x}'\in B_{\bm z'}(\varepsilon)$, we have
\begin{align*}
\lvert r_t^{m}(\bm{x},\bm{x}') - r(\bm{x},\bm{x}')\rvert \leq \lvert r_0^{m}(\bm{z},\bm{z}') - r(\bm{z},\bm{z}')\rvert+\lvert r(\bm{z},\bm{z}') - r(\bm{x},\bm{x}') \rvert&\\
+\lvert r_t^{m}(\bm{x},\bm{x}') -r_0^{m}(\bm{z},\bm{z}') \rvert
\end{align*}
Then, noticing that $r^m_t=\pt{r^{(1),m}_t+r^{(2),m}_t}/2$, we control the four terms on the right hand
side by Corollary \ref{prop:initialization of NTK}, Proposition \ref{prop:continuity of NTK} and \ref{prop:During training_uniform} 
respectively. We have shown that
\begin{align*}
 &\lvert r_0^{m}(\bm{z},\bm{z}') - r(\bm{z},\bm{z}') \rvert \leq O(m^{-0.2}),\qquad~\lvert r(\bm{x},\bm{x}') - r(\bm{z},\bm{z}')\rvert \leq O({1}/{m}),\\
 &\sup_{t\geq 0}\sup_{\bm{x}\in B_{\bm{z}}(\varepsilon)}\sup_{\bm{x}'\in B_{\bm{z}'}(\varepsilon)} |r_t^{m}(\bm{x},\bm{x}') - r_0^{m}(\bm{z},\bm{z}')| \leq O\left( m^{-{1}/{12}}\sqrt{\log m}\right),
\end{align*}
with probability at least $1-\delta/\pt{2|\mathcal{N}_{\varepsilon}|^2}$ if $m\geq C_1\log\mpt{C_2|\mathcal{N}_{\varepsilon}|^2/\delta}^5$ for some positive absolute constants $C_1>0$ and $C_2\geq 1$. There exists a polynomial $\poly_2(\cdot)$, such that when $m\geq\poly_2\mpt{\log(1/\delta)}$, we have $m\geq C_1\log\mpt{C_2|\mathcal{N}_{\varepsilon}|^2/\delta}^5$, since $\log |\mathcal{N}_{\varepsilon}| = O(\log m)$. By applying the union bound for any pair $\bm{z},\bm{z}'\in\mathcal{N}_\varepsilon$, we have with probability at least $1-\delta$, we have
\begin{align*}
 \sup_{t\geq 0}\sup_{\bm{x},\bm{x}'\in \mathcal{X}}\lvert r_{t}^{m}(\bm{x},\bm{x}') - r(\bm{x},\bm{x}') \rvert \leq O\mpt{m^{-{1}/{12}}\sqrt{\log m}}
\end{align*}
if $m\geq\poly_1\mpt{n,\norm{\bm{y}}_2,\lambda_0^{-1},\log(1/\delta)}+ \poly_2(\log(1/\delta))$.

\end{proof}

%% file: sb_Proof_generalization.tex
\section{Proofs in Section 2 and Section 4}

\begin{proof}[Proof of Lemma 2.1]
    For any $\bm{x},\bm{x}' \in \mathbb{S}^{d-1}$, we have the following decomposition of the NTK,
    \begin{align*}
        r(\bm{x},\bm{x}') = \sum_{k=0}^{\infty}\mu_k\sum_{h=1}^{N(d,k)}Y_{k,h}(\bm{x})Y_{k,h}(\bm{x}'),
    \end{align*}
where $Y_{k,h}$, $h=1,..,N(d,k)$ are spherical harmonic polynomials of degree $k$ and
\begin{align*}
    N(d,k)= \frac{\Gamma(k+d-2)}{\Gamma(d-1)\Gamma(k)}.
\end{align*}

By Stirling approximation, $\Gamma(x)=Cx^{x-1/2}\exp(-x)(1+O(\frac{1}{x}))$. Therefore, $N(d,k)$ is the order of $k^{d-2}$ and

\begin{align*}
    \lambda_j=\mu_l, \quad \sum_{i=1}^{l-1}N(d,2i)\leq j\leq \sum_{i=1}^{l}N(d,2i),
\end{align*}
i.e., $\lambda_j\asymp \mu_l$ for $j\in[(2l-2)^{d-1}, (2l)^{d-1}]$.
By Theorem 4.3 of \cite{belfer2021spectral}, the non-negative eigenvalues $\mu_k$ satisfy $\mu_k\asymp k^{-d}$, which implies $\lambda_j\asymp j^{-\frac{d}{d-1}}$.

\end{proof}

\begin{proof}[Proof of Theorem 4.1]

\begin{proposition}[Corollary 4.4 in \cite{lin2020optimal}]\label{prop:early:stopping}
    Suppose Assumption 1 holds. For any given $\delta\in(0,1)$, if the training process is stopped at $t_{*} \propto n^{\frac{d}{2d-1}}$ for the NTK regression, then for sufficiently large $n$, there exists a constant $C$ independent of $\delta$ and $n$, such that
    \begin{equation*}
        \mathcal{E}(f_{t_{*}}^{NTK})\leq Cn^{-\frac{d}{2d-1}}\log^{2}\frac{6}{\delta}
    \end{equation*}
    holds with probability at least $1-\delta$.
\end{proposition}

Setting $\epsilon=Cn^{-\frac{d}{2d-1}}\log^{2}\frac{6}{\delta}$ in Theorem 3.1 yields
\begin{equation*}
    \begin{aligned}
    \mathcal{E}(f_{t_*}^{m}) &\leq |\mathcal{E}(f_{t_*}^{m})-\mathcal{E}(f_{t_{*}}^{NTK})|+\mathcal{E}(f_{t_{*}}^{NTK}) \\
    &\leq 2Cn^{-\frac{d}{2d-1}}\log^{2}\frac{6}{\delta}.
    \end{aligned}
\end{equation*}

\end{proof}

%% file: main.bbl
\begin{thebibliography}{10}

\bibitem{allen2019convergence}
Zeyuan Allen-Zhu, Yuanzhi Li, and Zhao Song.
\newblock A convergence theory for deep learning via over-parameterization.
\newblock In {\em International Conference on Machine Learning}, pages
  242--252. PMLR, 2019.

\bibitem{beaglehole2022kernel}
Daniel Beaglehole, Mikhail Belkin, and Parthe Pandit.
\newblock Kernel {{Ridgeless Regression}} is {{Inconsistent}} in {{Low
  Dimensions}}, June 2022.

\bibitem{belfer2021spectral}
Yuval Belfer, Amnon Geifman, Meirav Galun, and Ronen Basri.
\newblock Spectral analysis of the neural tangent kernel for deep residual
  networks.
\newblock {\em arXiv preprint arXiv:2104.03093}, 2021.

\bibitem{blanchard2018optimal}
Gilles Blanchard and Nicole M{\"u}cke.
\newblock Optimal {{Rates}} for {{Regularization}} of {{Statistical Inverse
  Learning Problems}}.
\newblock {\em Foundations of Computational Mathematics}, 18(4):971--1013,
  August 2018.

\bibitem{bordelon2020spectrum}
Blake Bordelon, Abdulkadir Canatar, and Cengiz Pehlevan.
\newblock Spectrum dependent learning curves in kernel regression and wide
  neural networks.
\newblock In {\em International Conference on Machine Learning}, pages
  1024--1034. PMLR, 2020.

\bibitem{buchholz2022kernel}
Simon Buchholz.
\newblock Kernel interpolation in {{Sobolev}} spaces is not consistent in low
  dimensions.
\newblock In {\em Proceedings of {{Thirty Fifth Conference}} on {{Learning
  Theory}}}, pages 3410--3440. {PMLR}, June 2022.

\bibitem{caponnetto2007optimal}
Andrea Caponnetto and Ernesto De~Vito.
\newblock Optimal rates for the regularized least-squares algorithm.
\newblock {\em Foundations of Computational Mathematics}, 7(3):331--368, 2007.

\bibitem{chizat2019lazy}
L{\'e}na{\"i}c Chizat, Edouard Oyallon, and Francis Bach.
\newblock On {{Lazy Training}} in {{Differentiable Programming}}.
\newblock In {\em Advances in {{Neural Information Processing Systems}}},
  volume~32. {Curran Associates, Inc.}, 2019.

\bibitem{cui2021generalization}
Hugo Cui, Bruno Loureiro, Florent Krzakala, and Lenka Zdeborov{\'a}.
\newblock Generalization error rates in kernel regression: The crossover from
  the noiseless to noisy regime.
\newblock {\em Advances in Neural Information Processing Systems},
  34:10131--10143, 2021.

\bibitem{dasgupta2003elementary}
Sanjoy Dasgupta and Anupam Gupta.
\newblock An elementary proof of a theorem of johnson and lindenstrauss.
\newblock {\em Random Structures \& Algorithms}, 22(1):60--65, 2003.

\bibitem{du2019gradient}
Simon Du, Jason Lee, Haochuan Li, Liwei Wang, and Xiyu Zhai.
\newblock Gradient descent finds global minima of deep neural networks.
\newblock In {\em International conference on machine learning}, pages
  1675--1685. PMLR, 2019.

\bibitem{he2016deep}
Kaiming He, Xiangyu Zhang, Shaoqing Ren, and Jian Sun.
\newblock Deep residual learning for image recognition.
\newblock In {\em Proceedings of the IEEE conference on computer vision and
  pattern recognition}, pages 770--778, 2016.

\bibitem{hu2019simple}
Wei Hu, Zhiyuan Li, and Dingli Yu.
\newblock Simple and effective regularization methods for training on noisily
  labeled data with generalization guarantee.
\newblock {\em arXiv preprint arXiv:1905.11368}, 2019.

\bibitem{huang2020deep}
Kaixuan Huang, Yuqing Wang, Molei Tao, and Tuo Zhao.
\newblock Why do deep residual networks generalize better than deep feedforward
  networks?---a neural tangent kernel perspective.
\newblock {\em Advances in neural information processing systems},
  33:2698--2709, 2020.

\bibitem{jacot2018neural}
Arthur Jacot, Franck Gabriel, and Cl{\'e}ment Hongler.
\newblock Neural tangent kernel: Convergence and generalization in neural
  networks.
\newblock {\em arXiv preprint arXiv:1806.07572}, 2018.

\bibitem{jacot2020kernel}
Arthur Jacot, Berfin Simsek, Francesco Spadaro, Cl{\'e}ment Hongler, and Franck
  Gabriel.
\newblock Kernel alignment risk estimator: Risk prediction from training data.
\newblock {\em Advances in Neural Information Processing Systems},
  33:15568--15578, 2020.

\bibitem{krizhevsky2017imagenet}
Alex Krizhevsky, Ilya Sutskever, and Geoffrey~E Hinton.
\newblock Imagenet classification with deep convolutional neural networks.
\newblock {\em Communications of the ACM}, 60(6):84--90, 2017.

\bibitem{lai2023generalization}
Jianfa Lai, Manyun Xu, Rui Chen, and Qian Lin.
\newblock Generalization ability of wide neural networks on $\mathbb {R}$.
\newblock {\em arXiv preprint arXiv:2302.05933}, 2023.

\bibitem{lee2019wide}
Jaehoon Lee, Lechao Xiao, Samuel Schoenholz, Yasaman Bahri, Roman Novak, Jascha
  Sohl-Dickstein, and Jeffrey Pennington.
\newblock Wide neural networks of any depth evolve as linear models under
  gradient descent.
\newblock {\em Advances in neural information processing systems},
  32:8572--8583, 2019.

\bibitem{li2018visualizing}
Hao Li, Zheng Xu, Gavin Taylor, Christoph Studer, and Tom Goldstein.
\newblock Visualizing the loss landscape of neural nets.
\newblock {\em Advances in neural information processing systems}, 31, 2018.

\bibitem{li2023Statistical}
Yicheng Li, Zixiong Yu, Guhan Chen, and Qian Lin.
\newblock Statistical optimality of deep wide neural networks.
\newblock {\em arXiv preprint arXiv:2305.02657}, 2023.

\bibitem{li2023kernel}
Yicheng Li, Haobo Zhang, and Qian Lin.
\newblock Kernel interpolation generalizes poorly.
\newblock {\em arXiv preprint arXiv:2303.15809}, 2023.

\bibitem{liang2020just}
Tengyuan Liang and Alexander Rakhlin.
\newblock Just interpolate: Kernel “ridgeless” regression can generalize.
\newblock {\em The Annals of Statistics}, 48(3):1329--1347, 2020.

\bibitem{lin2020optimal}
Junhong Lin, Alessandro Rudi, Lorenzo Rosasco, and Volkan Cevher.
\newblock Optimal rates for spectral algorithms with least-squares regression
  over {{Hilbert}} spaces.
\newblock {\em Applied and Computational Harmonic Analysis}, 48(3):868--890,
  May 2020.

\bibitem{liu2019towards}
Tianyi Liu, Minshuo Chen, Mo~Zhou, Simon~S Du, Enlu Zhou, and Tuo Zhao.
\newblock Towards understanding the importance of shortcut connections in
  residual networks.
\newblock {\em Advances in neural information processing systems}, 32, 2019.

\bibitem{ma2018priori}
Chao Ma, Lei Wu, et~al.
\newblock A priori estimates of the population risk for two-layer neural
  networks.
\newblock {\em arXiv preprint arXiv:1810.06397}, 2018.

\bibitem{ma2022barron}
Chao Ma, Lei Wu, et~al.
\newblock The barron space and the flow-induced function spaces for neural
  network models.
\newblock {\em Constructive Approximation}, 55(1):369--406, 2022.

\bibitem{mallinar2022benign}
Neil Mallinar, James~B Simon, Amirhesam Abedsoltan, Parthe Pandit, Mikhail
  Belkin, and Preetum Nakkiran.
\newblock Benign, tempered, or catastrophic: A taxonomy of overfitting.
\newblock {\em arXiv preprint arXiv:2207.06569}, 2022.

\bibitem{nakkiran2021deep}
Preetum Nakkiran, Gal Kaplun, Yamini Bansal, Tristan Yang, Boaz Barak, and Ilya
  Sutskever.
\newblock Deep double descent: Where bigger models and more data hurt.
\newblock {\em Journal of Statistical Mechanics: Theory and Experiment},
  2021(12):124003, 2021.

\bibitem{rakhlin2019consistency}
Alexander Rakhlin and Xiyu Zhai.
\newblock Consistency of {{Interpolation}} with {{Laplace Kernels}} is a
  {{High-Dimensional Phenomenon}}.
\newblock In {\em Proceedings of the {{Thirty-Second Conference}} on {{Learning
  Theory}}}, pages 2595--2623. {PMLR}, June 2019.

\bibitem{raskutti2014early}
Garvesh Raskutti, Martin~J Wainwright, and Bin Yu.
\newblock Early stopping and non-parametric regression: an optimal
  data-dependent stopping rule.
\newblock {\em The Journal of Machine Learning Research}, 15(1):335--366, 2014.

\bibitem{tirer2022kernel}
Tom Tirer, Joan Bruna, and Raja Giryes.
\newblock Kernel-based smoothness analysis of residual networks.
\newblock In {\em Mathematical and Scientific Machine Learning}, pages
  921--954. PMLR, 2022.

\bibitem{vaswani2017attention}
Ashish Vaswani, Noam Shazeer, Niki Parmar, Jakob Uszkoreit, Llion Jones,
  Aidan~N Gomez, {\L}ukasz Kaiser, and Illia Polosukhin.
\newblock Attention is all you need.
\newblock {\em Advances in neural information processing systems}, 30, 2017.

\bibitem{veit2016residual}
Andreas Veit, Michael~J Wilber, and Serge Belongie.
\newblock Residual networks behave like ensembles of relatively shallow
  networks.
\newblock {\em Advances in neural information processing systems}, 29, 2016.

\bibitem{vershynin2010introduction}
Roman Vershynin.
\newblock Introduction to the non-asymptotic analysis of random matrices.
\newblock {\em arXiv preprint arXiv:1011.3027}, 2010.

\bibitem{ma2020rademacher}
E~Weinan, Chao Ma, Qingcan Wang, et~al.
\newblock Rademacher complexity and the generalization error of residual
  networks.
\newblock {\em Communications in Mathematical Sciences}, 18(6):1755--1774,
  2020.

\bibitem{ma2019generalization}
E~Weinan, Chao Ma, Lei Wu, et~al.
\newblock The generalization error of the minimum-norm solutions for
  over-parameterized neural networks.
\newblock {\em arXiv preprint arXiv:1912.06987}, 2019.

\bibitem{yao2007early}
Yuan Yao, Lorenzo Rosasco, and Andrea Caponnetto.
\newblock On early stopping in gradient descent learning.
\newblock {\em Constructive Approximation}, 26(2):289--315, 2007.

\bibitem{zhang2016understanding}
Chiyuan Zhang, Samy Bengio, Moritz Hardt, Benjamin Recht, and Oriol Vinyals.
\newblock Understanding deep learning requires rethinking generalization.
\newblock In {\em International {{Conference}} on {{Learning
  Representations}}}, 2016.

\bibitem{zhang2023optimality}
Haobo Zhang, Yicheng Li, and Qian Lin.
\newblock On the optimality of misspecified spectral algorithms.
\newblock {\em arXiv preprint arXiv:2303.14942}, 2023.

\end{thebibliography}
